\documentclass[a4paper,fleqn]{cas-dc}

\usepackage[authoryear,longnamesfirst]{natbib}
\usepackage{subfigure}
\usepackage{xcolor}
\usepackage{soul}

\def\tsc#1{\csdef{#1}{\textsc{\lowercase{#1}}\xspace}}
\tsc{WGM}
\tsc{QE}
\tsc{EP}
\tsc{PMS}
\tsc{BEC}
\tsc{DE}
\begin{document}
\let\WriteBookmarks\relax
\def\floatpagepagefraction{1}
\def\textpagefraction{.001}

% Short title
\shorttitle{DyFFPAD}

% Short author
\shortauthors{A. Rai et~al.}

% Main title of the paper
\title [mode = title]{DyFFPAD: Dynamic Fusion of Convolutional and Handcrafted Features for Fingerprint Presentation Attack Detection}                      
% Title footnote mark
% eg: \tnotemark[1]
\tnotemark[1,2]

% Title footnote 1.
% eg: \tnotetext[1]{Title footnote text}
% \tnotetext[<tnote number>]{<tnote text>} 
% \tnotetext[1]{This document is the results of the research
%    project funded by the National Science Foundation.}

% \tnotetext[2]{The second title footnote which is a longer text matter
%    to fill through the whole text width and overflow into
%    another line in the footnotes area of the first page.}

% First author
%
% Options: Use if required
% eg: \author[1,3]{Author Name}[type=editor,
%       style=chinese,
%       auid=000,
%       bioid=1,
%       prefix=Sir,
%       orcid=0000-0000-0000-0000,
%       facebook=<facebook id>,
%       twitter=<twitter id>,
%       linkedin=<linkedin id>,
%       gplus=<gplus id>]
\author[1]{Anuj Rai}[type=editor,
                        auid=000,bioid=1,
                        prefix=,
                        role=,
                        orcid=]

\ead{anujrai.iiti@gmail.com}
\author[1]{Prasheel Kumar Tiwari}[type=editor,
                        auid=000,bioid=1,
                        prefix=,
                        role=,
                        orcid=]
\ead{prashilkumartiwari@gmail.com}
\author[1]{Jyotishna Baishya}[type=editor,
                        auid=000,bioid=1,
                        prefix=,
                        role=,
                        orcid=]
\ead{jyotishnab800@gmail.com}
\author[2]{Ramprakash Sharma}[type=editor,
                        auid=000,bioid=1,
                        prefix=,
                        role=,
                        orcid=]
\ead{rampsharma28@gmail.com}
\author[1]{Somnath Dey}[type=editor,
                        auid=000,bioid=1,
                        prefix=,
                        role=,
                        orcid=]
\ead{somnathd@iiti.ac.in}
% Corresponding author indication
%\cormark[1]

% Footnote of the first author
%\fnmark[1]

% Email id of the first author
% \ead{cvr_1@tug.org.in}

% % URL of the first author
% \ead[url]{www.cvr.cc, cvr@sayahna.org}

%  Credit authorship
%\credit{Conceptualization of this study, Methodology, Software}

% Address/affiliation
\affiliation[1]{organization={IIT Indore},
    addressline={Department of Computer Science and Engineering, Indian Institute of Technology Indore}, 
    city={Indore},
    % citysep={}, % Uncomment if no comma needed between city and postcode
    postcode={453552}, 
    state={Madhya Pradesh},
    country={India}}

\affiliation[2]{organization={NIT Hamirpur},
    addressline={Department of Computer Science and Engineering, National Institute of Technology Hamirpur}, 
    city={Hamirpur},
    % citysep={}, % Uncomment if no comma needed between city and postcode
    postcode={177005}, 
    state={Himanchal Pradesh},
    country={India}}

% \fnmark[2]
% \ead{cvr3@sayahna.org}
% \ead[URL]{www.sayahna.org}

%\credit{Data curation, Writing - Original draft preparation}

% Address/affiliation

% Corresponding author text
% \cortext[cor1]{Corresponding author}
% \cortext[cor2]{Principal corresponding author}

% % Footnote text
% \fntext[fn1]{This is the first author footnote. but is common to third
%   author as well.}
% \fntext[fn2]{Another author footnote, this is a very long footnote and
%   it should be a really long footnote. But this footnote is not yet
%   sufficiently long enough to make two lines of footnote text.}

% % For a title note without a number/mark
% \nonumnote{This note has no numbers. In this work we demonstrate $a_b$
%   the formation Y\_1 of a new type of polariton on the interface
%   between a cuprous oxide slab and a polystyrene micro-sphere placed
%   on the slab.
%   }

% Here goes the abstract
\begin{abstract}
Automatic fingerprint recognition systems suffer from the threat of presentation attacks due to their wide range of deployment in areas including national borders and commercial applications. A presentation attack can be performed by creating a spoof of a user's fingerprint with or without their consent. This paper presents a deep neural network-based solution to safeguard fingerprint recognition systems against such attacks. The proposed dynamic ensemble of deep CNN and handcrafted features to detect presentation attacks in known-material and unknown-material scenarios of the liveness detection. The proposed presentation attack detection model, in this way, utilizes the capabilities of both deep CNN and handcrafted features techniques and exhibits better performance than their individual performances. We have validated our proposed method on benchmark databases from the Liveness Detection Competition in 2015, 2017, and 2019, yielding overall accuracy of 96.10\%, 96.49\%, and 94.99\% on them, respectively. The proposed method outperforms state-of-the-art methods in terms of classification accuracy. 
\end{abstract}

\begin{highlights}
\item  This paper presents a Deep Neural Network-based method to protect the Automatic Fingerprint Recognition System (AFRS) from Presentation Attacks (PAs).
\item  It presents a novel fusion of convolutional and handcrafted features which address the limitations that occur in the static fusion of these features.
\item The proposed method performs better than state-of-the-art methods in benchmark scenarios.
\end{highlights}


\maketitle

\section{Introduction}
{A}{utomatic} Fingerprint Recognition Systems (AFRS) are most widely used for the purpose of person authentication and verification \cite{Finger_NN}. Their user-friendliness and cost-effectiveness make them popular while validating the identity of persons at airports, national borders, and distribution of government-funded aid. The utilization of these systems in a wide range of applications, also makes them vulnerable to some internal and external security threats. Presentation Attack (PA), or spoofing, is an external attack where an artificial artifact of a user's finger is presented to the sensor of an Automated Fingerprint Recognition System (AFRS). PAs or spoofs can be created either by a cooperative method or a non-cooperative method of spoofing. Fabrication materials such as latex, woodglue, gelatine, silicon, ecoflex, etc. are available at a very low cost to create the spoof of a fingerprint. Fingerprint Presentation Attack Detection (FPAD) serves as a countermeasure to these attacks and to empower an AFRS against them. The recent FPAD methods suggested by various researchers are categorized into two broad classes including hardware-based and software-based. The methods based on hardware are quite expensive due to the utilization of additional sensing devices that measure natural properties such as heart rate, odor, temperature, etc. The utilization of these sensors makes hardware-based methods less user-friendly and quite expensive for an organization to use in access-control applications. On the other side, software-based methods require only fingerprint samples which makes them friendly to the user and cost-effective to the organization as compared to hardware-based methods.  
Current software-based methods can be further categorized into perspiration and pore based-methods \cite{b16_espinoza}, \cite{b19_abhyankar} statistical and handcrafted feature-based methods \cite{b5_sharma1}, \cite{b37_deepika}, \cite{xia_2}, \cite{Anuj_2} and deep-learning based methods \cite{b14_chugh1}, \cite{b15_chugh2}, \cite{anuj_1}, and \cite{anuj_4}. Perspiration-based methods suffer from dryness in the environment which sometimes causes the rejection of a genuine fingerprint sample. Similarly, pore-based methods necessitate the sensing device to capture a high-definition image of the fingertip. This necessity impacts the overall cost of these methods. Existing works utilizing statistical and handcrafted feature extract some predefined features from the input fingerprint sample to classify them as live or spoof. These methods are affected by the quality of the fingerprint samples. However, some of these methods including \cite{b5_sharma1}, \cite{b37_deepika} have shown good FPAD capability while the spoof sample is created using known material but they have not been tested properly on the spoofs fabricated with unknown materials. Deep learning methods consist of a classifier that possesses convolutional layers as a feature extractor. These layers have an unmatched capability of extracting minute features from the input samples. In recent times, various deep learning methods have been suggested by various researchers which have exhibited better performance than other software-based methods but they also suffer while being tested with the spoofs fabricated with unknown materials. 

\begin{figure*}[t]
	\centering
    \resizebox{0.8\textwidth}{!}{
	\subfigure[LivDet 2015 Biometrika \hspace{42 mm} (c) LivDet 2015 Crossmatch]
	{	
		\includegraphics[width=0.40\textwidth]{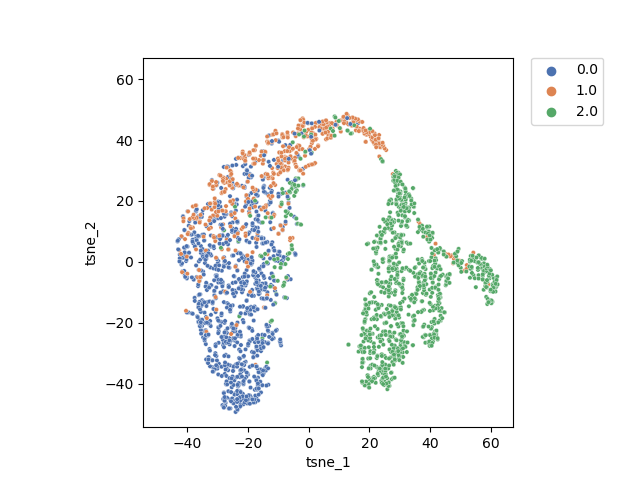}
		\hspace{2 mm}
		\includegraphics[width=0.40\textwidth ]{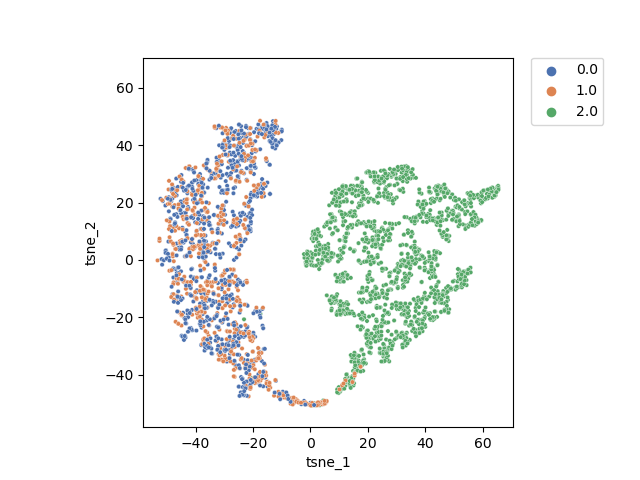}
	}}
	\vspace{4 mm}
	\resizebox{0.8\textwidth}{!}{
	\subfigure[LivDet 2015 Digital Persona \hspace{40 mm} (d) LivDet 2015 Greenbit]
	{
		\includegraphics[width=0.40\textwidth]{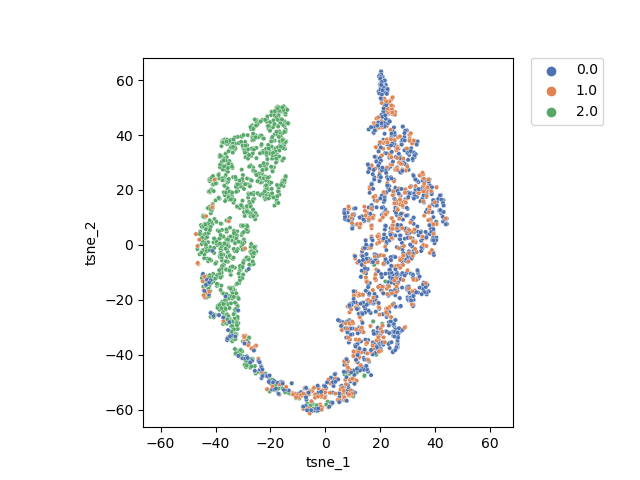}
		\hspace{2 mm}
		\includegraphics[width=0.40\textwidth]{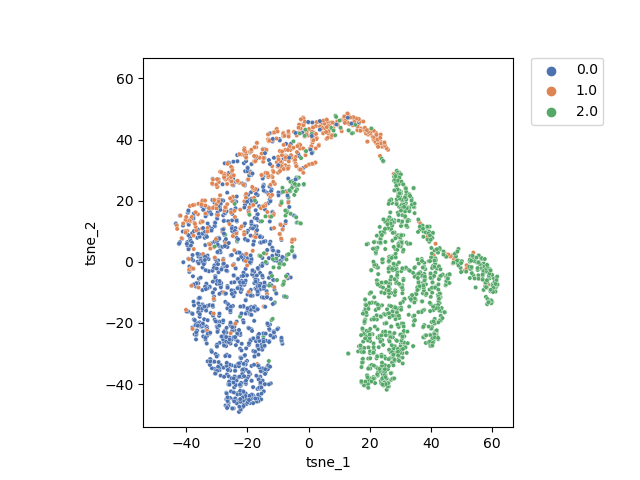}
	}}
	\vspace{4mm}
	%\vspace{-4 mm}
	\caption{2D - tsne plots for the features generated by the proposed DyFFPAD model on LivDet 2015 (a) Biometrika, (b) Digital Persona, (c) Crossmatch, and (d) Greenbit sensors. 0 represents spoofs fabricated with known materials, 1 represents spoofs fabricated with unknown materials, and 2 represents live fingerprint samples. It depicts the proposed model's ability for generalization across both known and unknown materials in fabricated spoofs, effectively distinguishing them from live fingerprint samples.}
	\label{TSNE}
	\vspace{-2mm}
\end{figure*}

In our proposed work, a dynamic ensemble of handcrafted and deep features-based model is presented. This \textbf{D}ynamic \textbf{F}usion of convolutional and handcrafted features for \textbf{F}ingerprint \textbf{P}resentation \textbf{A}ttack \textbf{D}etection \textbf{(DyFFPAD)}, is an end-to-end approach that dynamically incorporates deep and handcrafted features. The model is build using two sub-models that work in collaboration to classify the live and spoof fingerprint samples. The first sub-model is a feature extractor model comprising a Deep Neural Network (DNN) which is empowered with an image descriptor namely Local Phase Quantization (LPQ) and a set of quality-based handcrafted features including Ridge Valley Clarity (RVC), Gabor quality, Frequency Domain Analysis (FDA), and Orientation Flow (OFL), etc. On the other hand, the second sub-model is a DenseNet architecture's feature base attached to a custom-made DNN. DenseNet has shown remarkable performance while being tested on some well-known image classification problem databases including MNIST \cite{mnist}, CIFAR \cite{cifar}, and imagenet \cite{imagenet}. The output of both the sub-models is combined and fed to another DNN which has a single neuron in the last for the binary classification. The proposed model addresses the issue of developing a dynamic fusion of convolutional and handcrafted features in a better way as it does not require any intermediary stage of data preparation and training of multiple classifiers, unlike other static fusion-based models presented in \citep{Nguyen,Nguyen2, Daksha_Yadav}. The TSNE plots depicted in Fig. $\ref{TSNE}$ show that the proposed method can distinguish between live samples, and spoof samples irrespective of their fabrication materials. The proposed model shows the same performance on different datasets and its efficacy is not affected by known or unknown materials used for spoof fabrication. The proposed method is evaluated using Liveness Detection Competition (LivDet) databases, demonstrating superior results compared to state-of-the-art methods for the evaluation protocol of intra-sensor same-material and intra-sensor cross-material. A detailed comparison with the state-of-the-art methods indicates the supremacy of the proposed method while detecting the PAs in benchmark protocols of FPAD. 

The major contributions of our proposed work are highlighted below. 

1. A novel end-to-end architecture is proposed that embodies the capabilities of convolutional filters along with LPQ and handcrafted features to detect PAs fabricated using known and unknown materials.

2. A dynamic fusion of handcrafted and convolutional features is proposed which addresses the limitations that occur in static fusion of these features. 

3. An experimental study is presented that shows the performance of PAs on the DenseNet classifier and DNNs which are trained with handcrafted and LPQ features individually and a combined feature vector comprised of LPQ and handcrafted feature values.

4. An exhaustive comparative study, conducted on intra-sensor known-material and unknown-material protocols, reveals that the proposed method outperforms others in terms of classification accuracy.

5. From TSNE representations, it is evident that the proposed model demonstrates consistency while detecting spoofs fabricated with known and unknown materials.
\\

The rest of this paper is arranged in following order: Section \ref{related work} presents related work by researchers for the detection of PAs, highlighting their advantages and limitations. Section \ref{proposed work} provides the details of architecture and design of proposed method.  Section \ref{experimentalresults} presents the experimental setup and protocol details while Section \ref{EXPCA} provides the experimental results, followed by a comparative analysis. Finally, the conclusion is provided in Section \ref{conclusion}.

\section{Related Work} \label{related work}
PAs poses a significant security threat to the AFRS which has brought the attention of researchers toward the development of cost-effective and user-friendly solutions to this problem. We have provided an overview of some of the prominent methods put forth by researchers to keep the AFRS safe against PAs in this section. Existing methods utilizes perspiration and pore features, statistical and handcrafted features, and deep learning features. The methods that fall into these categories and their limitations are discussed in this section. 

\subsection{Perspiration and pore based-methods}
Perspiration is a natural phenomenon in human skin that is caused by the presence of tiny holes or pores. As spoofs created with different materials do not have this natural property, it could be utilized to develop a method that distinguishes between live and spoof fingerprints based on this property. In this attempt, \cite{Derakshini2003} leveraged the presence of perspiration and diffusion patterns of sweat to differentiate between live and spoof fingerprints. Further, \cite{b19_abhyankar} have proposed a wavelet-based approach for the detection of PAs using the fingerprint's perspiration property. They have used a custom-made fingerprint database to evaluate the performance of their proposed method. The count of pores within a live fingerprint and its counterfeits crafted from various materials can exhibit variability as it is difficult to reflect pores in the spoofs fingerprints. \cite{b16_espinoza} utilized this dissimilarity as a distinguishing feature to detect PAs on a custom-made dataset. \cite{b17_marcialis} also proposed a similar method that detects pores from multiple fingerprints captured at an interval of five seconds. The number of pores in the subsequent samples is further utilized to detect PAs on a their custom-made database having 8,960 live and 7,760 spoof samples. \cite{Manivanan2010} utilized high-pass filtering followed by correlation filtering techniques \cite{Memon2011} to identify the sweat pores present on fingerprint ridges. Their results demonstrate distinguishable correlation peaks corresponding to the sweat pores on the ridges.
Regardless of how the perspiration pattern is employed to detect PAs, its presence is dependent on the temperature of the surrounding environment. Sometimes, even a live finger does not exhibit this property in a dry environment which occasionally prompts the FPAD method to mis-classify this live sample as a spoof. Similarly, the pore-based methods cost high due to the utilization of a high-resolution sensing device that is capable of collecting fingerprint samples with higher dpi ($>=$1000 pixels per inch). These factors collectively make perspiration-based and pore-based methods less economical and less convenient to use.

\subsection{Handcrafted and statistical feature based-methods}
The inherent disparities between the natural attributes of a finger's skin and counterfeit spoofs, such as variations in color and moisture levels, manifest in the quality of collected samples. This phenomenon prompted researchers to utilize the quality of captured fingerprint samples as a distinguishing feature. This section describes some of the approaches that fall within this category. \cite{b11} employed a range of statistical features, encompassing deviation, hyper-flatness, average brightness, standard deviation, skewness, variance, hyper-skewness, kurtosis, and differential image features, enabling detection of live and spoof samples using Support Vector Machine (SVM) classifier. The efficacy of the proposed approach is evaluated through experimentation on the ATVSFFp database, comprising 272 genuine and 270 spoof fingerprint samples. In a similar attempt, \cite{xia_2} developed an image descriptor that captures both gradient characteristics and intensity variance, resulting in a composite feature vector that is fed to an SVM classifier to identify spoofs and live fingerprints The proposed method is validated on LivDet 2011, 2013, and 2015 datasets. \cite{b23_kim} have proposed another image descriptor based on the local coherence property of fingerprints to train an SVM classifier on LivDet 2009, 2011, 2013, and 2015 and ATVSFFp databases. \cite{b44_yuan3} introduced a novel approach that utilizes gradient properties for image analysis. This method involves generating image gradient matrices through diverse quantization operators which are derived using the Laplacian operator. These computed matrices are then employed to train a back-propagation neural network using the LivDet 2013 database. \cite{Gragnaniello2013} employed the Weber Local Descriptor (WLD) to extract gradient information and digital excitation from fingerprint samples which are subsequently used to train an SVM classifier. They have evaluated their method using the LivDet 2009 and 2011 databases. \cite{Gragnaniello2015} proposed a novel image descriptor named Local Contrast Phase Descriptor (LCPD) which is a spatial domain and rotation invariant version of Local Phase Quantization (LPQ). The proposed method is validated on LivDet 2011 database. \cite{Rattani2014_1} proposed an incremental learning based approach that consists of a novel material detector along with the classifier. This detector divides the input fingerprint samples into three categories: live, spoof, and unknown. Samples identified as 'novel' are then utilized to train the classifier and evaluated on LivDet 2011 database. In continuation to this work, \cite{Rattani2015} proposed a novel approach that employs a Weibull-calibrated SVM for the classification of fingerprint samples. This approach integrates one-class and binary SVMs, demonstrating a significant improvement over their previous method \cite{Rattani2014_1}. Since the finger skin and spoof fabrication materials exhibit different levels of elasticity the resultant fingerprints have some differences in terms of quality. The difference in the live and spoof samples fabricated with different materials results in uneven ridge and valley widths and the fabrication process introduces other quality irregularities in the spoofs. In an attempt to develop a quality-based FPAD model, \cite{b5_sharma1} utilized some fingerprint quality features for the detection of PAs. They extracted Ridge and Valley Smoothness (RVS), Ridge and Valley Clarity (RVC),  Orientation Certainty Level (OCL), and Frequency Domain Analysis (FDA) to constitute a feature vector for training of Random-Forest (RF) classifier. They have carried out experimental evaluations on LivDet 2009, 2011, 2013, and 2015 databases. Similarly, \cite{b37_deepika} proposed a Local Adaptive Binary Pattern (LABP) image descriptor which is a novel variant of the Local Binary Pattern (LBP). The LABP features are combined with the existing Complete Local Binary Pattern (CLBP) and Binary Statistical Image Features (BSIF) to train both an SVM classifier and a Deep Neural Network (DNN). The efficacy of the proposed method is validated across multiple databases, including LivDet 2009, 2011, 2013, and 2015. \cite{Rubab} proposed a novel method that focuses on liveness features in input fingerprint samples. In this method, the input sample after pre-processing undergoes extraction of two-channel information. The first channel consists of ridge contour in the frequency domain while the second channel consists of ridge contour in the spatial domain. It also extracts minutiae information along with liveness features using the proposed Comprehensive Local Phase Quantization (CLPQ) image descriptor. All the features are used to train an SVM. The proposed method is validated on LivDet 2011, 2013, and 2015 databases.  The quality of the fingerprint samples plays a vital role in PA detection by these methods. Some of the methods including \cite{b5_sharma1}, \cite{b37_deepika} \cite{b23_kim} exhibit good performance while being evaluated in the known material scenario.

\subsection{Deep learning-based-methods} Deep Convolutional Neural Networks (CNNs) excel at extracting minute features from input image samples. These models have shown remarkable classification performance on image datasets such as MNIST \cite{mnist}, CIFAR \cite{cifar}, and ImageNet \cite{imagenet}. This ability of the CNNs has prompted researchers to explore their application in PAD. \cite{Park2016} extracted multiple patches from segmented fingerprints and trained a custom-made CNN composed of three convolutional and one fully connected layer. Their method utilizes a voting strategy instead of score-level fusion to compute the overall liveness score of fingerprint samples in LivDet 2009 database. In a similar attempt, \cite{Toosi2017} applied fingerprint foreground segmentation on fingerprint images followed by patch extraction. The extracted patches are further used to train a pre-trained Alexnet \cite{Alexnet2012} CNN classifier using LivDet databases. \cite{Nogueira2014} trained two separate SVMs using the features extracted with Local Binary Pattern (LBP) image descriptor and a single-layer CNN respectively. They also incorporated various data augmentation techniques including image rotation and scaling. The limitation of this work is the feature extraction and the classification tasks are performed as separate tasks which do not allow the entire system to learn their parameters together. In another work, \cite{Nogueira2016} utilized pre-trained CNN (VGG, Alexnet, and CNN) along with an SVC classifier for handcrafted features and applied score-level fusion for the detection of PAs on LivDet databases. \cite{b4_uliyan} proposed a method that utilizes a Deep Boltzmann Machine (DBM) for feature extraction and a Restricted Boltzmann Machine (RBM) to establish relationships between these features. \cite{Yuan2022} have presented a multi-modal approach utilizing multiple pre-trained CNNs to detect the PAs on LivDet 2011, 2013, 2015, and NUAA face databases.
\cite{b14_chugh1} developed a novel method that crops the patches centered at minutiae points for training the MobileNet classifier. The average liveness score of all the patches is computed to calculate the global liveness score for an input sample. They have tested their method using Michigan State University's (MSU) FPAD database, and LivDet 2011, 2013, 2015. 
 \cite{b31_zhang} proposed a novel CNN architecture with a series of improved residual connected blocks. It reduces the problem of over-fitting and also takes less processing time. 
 \cite{GONZÁLEZ-SOLER} proposed a technique for FPAD in the cross-material scenario that leverages three representations that integrate both global and local information: Bag-of-Words (BoW), Vector Locally Aggregated Descriptors (VLAD), and Fisher Vector (FV). The dense-SIFT descriptor is computed at multiple scales, and the extracted features are encoded using these three feature encoding methods. The encoded features are then utilized to train an SVM classifier. The proposed method is validated on LivDet 2011, 2013, 2015, 2017, and 2019 databases with the FV encoding scheme demonstrating superior performance compared to others. \cite{JingLi} proposed an FPAD method that utilizes the Fisher Vector (FV) method for representing images by aggregating local image features into a single descriptor. The proposed method first transforms the image into local and global feature domains and then extracts discriminating features from them. The local frequency features, local spatial features, and global frequency features are then utilized to classify between live and spoof samples. The proposed method is validated on LivDet 2011, 2013, and 2015 databases.
Since, novel fabrication materials are being discovered nowadays, generalizing an open-set FPAD model is a challenging issue. \cite{b15_chugh2}, in continuation to their previous work \cite{b14_chugh1}, proposed an approach for detecting spoofs fabricated with unknown materials. They augmented the fingerprint databases with synthesized fingerprint patches representing the unknown materials spoofs for training the MobileNet classifier. The proposed method is validated on LivDet 2017, ATVSFFp, and MSU-FPAD databases. Generative Adversarial Networks can also be utilized to augment the training database with realistic spoof samples. In a similar attempt, \cite{anuj_1} utilized Wasserstein GAN to generate fingerprint patches to train the DenseNet classifier against the spoofs fabricated with unknown materials. The proposed method is validated using LivDet 2015, 2017, and 2019 databases. In another work, \cite{anuj_4} proposed a novel ensemble of MobileNet and SVC for the detection of PAs in intra-sensor and cross-sensor scenarios of FPAD. The proposed method utilizes MobileNet's convolutional base for feature extraction and SVC as a classifier. The proposed method is validated on LivDet 2011, 2013, 2015, 2017, and 2019 databases.
\\Although the Deep learning-based methods are quite effective for image classification problems still, they suffer from low classification accuracy in the area of FPAD while being tested in the cross-material protocol of FPAD. 
The methods presented in this sub-section prove that the deep learning-based methods are superior as compared with the methods of other categories but their performance is required to be improved in cross-material scenario of FPAD. This objective can be attained by developing a hybrid model that utilizes the power of both convolutional and handcrafted features. The proposed model should also be efficient enough to classify the input sample in a minimum amount of time to be incorporated with the AFRS. 

\section{Proposed Work \label{proposed work}}
In this paper, an ensemble of deep and handcrafted features is proposed for the detection of spoofs in different evaluation protocols of FPAD. It consists of two modules that collaborate to detect the PAs. The first module is a DNN that is fed with the Local Phase Quantization (LPQ) features and existing handcrafted features. The second module is the DenseNet CNN classifier which has shown remarkable performance while being validated on various image databases including imagenet \cite{imagenet}. Both of the modules are fused dynamically which results in better learning of them as compared with their individual performances. The details of LPQ, handcrafted features, DenseNet, and DyFFPAD are provided in the following sub-sections.

\subsection{Local Phase Quantization (LPQ) \label{LPQ1}} LPQ \cite{LPQ} works on the Fourier transform's blur-insensitive property which makes it robust against the redundant information and blur present in the input sample. We have utilized it as a prominent feature due to its capability of exploiting the minute information that gets missing in the fabrication process of a spoof. The formulation of the LPQ descriptor is denoted as Eq. \ref{LPQ}.
\begin{equation}
\label{LPQ}
  f_{x}(u) = \sum_{} f(y)w(y-x)e^{-j2\pi uy}
\end{equation}
Here, $f()$ corresponds to the output of the short-term Fourier transform, $f_{x}$ represents the local Fourier coefficients at four distinct frequency values, and $w()$ defines the neighborhood as a window function.
\subsection{Handcrafted Features \label{handc}} The fabrication process impacts the quality of the spoofs fabricated with various materials. According to recent research \cite{b5_sharma1} \cite{b37_deepika}, handcrafted features are proven to be useful while detecting PAs. In this work, we have utilized some existing handcrafted features \cite{b5_sharma1}, \cite{olsen} including ridge-valley clarity, gabor quality,  ridge-valley smoothness, number of abnormal ridges and valleys, orientation certainty level, and frequency domain analysis to estimate the quality of the input fingerprint sample. The details of these features is provided in the following subsections.

\subsubsection{Ridge and Valley Smoothness (RVS and VWS)} It denotes the smoothness of ridge width and valley width which is exhibited by the live sample but not by the spoof. This irregularity in the spoofs is caused because of the varying elasticity levels of spoof fabrication materials. These materials do not possess the elasticity same as human skin. In addition to this, the pressure applied on the fingertip at the time of sample collection is also a potential reason behind this. RWS and VWS are computed block-wise by cropping the vertical ridge-valley structure after the rotation and binarization. Thereafter, pixels of the block are labeled as ridge or valley with the help of a linear regression algorithm. For each horizontal line of the block, ridge and valley width is computed. RWS and VWS features are obtained by averaging the standard deviation of ridge and valley widths.

\subsubsection{Frequency Domain Analysis (FDA)}
FDA of a local block is extracted by computing the ridge-valley structure's 1D signature. The frequency of the sinusoidal ridge-valley structure is calculated using the discrete Fourier transform of this 1D signature. FDA is consistent in live fingerprints while it is having a varying frequency of sinusoidal ridge-valley patterns in spoof fingerprints. 

\subsubsection{Orientation Certainty Level (OCL)} 
OCL is the block-wise intensity of the energy concentration along the prominent ridge direction. The gradient vector is used to calculate the ratio of two eigenvalues of the covariance matrix. 

\subsubsection{Gabor Quality (G)}
 To calculate gabor quality at the local level, filter bank with many orientation on each pixel is applied. For a small number of filters with similar orientations as of block, the Gabor response with a normal ridge-valley pattern remains high, but it is low and steady for the same block with an improper ridge-valley structure. Lastly, the standard deviation of the gabor filter bank's output is used to determine the Gabor quality (G) of a block. Gabor filter is well-suited for texture analysis, feature extraction, edge detection, etc.

\subsubsection{Ridge-Valley Clarity (RVC) }
The ridge and valleys present in live fingerprints block are found to be separated consistently. This happens due to the uniform elasticity property of human finger skin.This separation, on the other hand, differs in spoof samples due to inappropriate widths of ridges and valleys induced by the different elasticity levels of spoofing materials employed in spoof fabrication. The RVC feature is computed by dividing the original sample into fixed-sized blocks and then computing the average widths of ridge and valley in the block. 

\subsubsection{Abnormal ridges and valleys ($R_{ab}$ and $V_{ab}$)}
The width of ridges and valleys is found to be in a range of 5 to 10 pixels in samples of human fingers captured from a sensing device with a resolution of 500 dpi. Since the human skin and spoofing materials have different levels of elasticity, the width of ridges and valleys differs in corresponding samples. An abnormality of ridge or valley is identified if its width divergence exceeds a particular threshold in various rows of the block. 

\subsubsection{Feature vector from quality features}
The final feature vector representing the collection of all above discussed features is represented as Eq. (\ref{Feature_vector}) is generated using the mean ($\mu$) and standard deviation ($\sigma$) of the features discussed above.

\begin{equation}
\begin{aligned}
Q = 
\{RWS^{\mu}, RWS^{\sigma}, VWS^{\mu}, VWS^{\sigma}, R_{ab}^{\mu},
 V_{ab}^{\mu}, RVC^{\mu}, \\ RVC^{\sigma}, FDA^{\mu}, FDA^{\sigma}, OCL^{\mu}, OCL^{\sigma}, Gabor^{\mu}\}
\end{aligned}
\label{Feature_vector}
\end{equation}
\subsection{DenseNet-121}
\begin{figure*}[]
	\centering
	\resizebox{1.0\textwidth}{!}{
		{
			
			\includegraphics[]{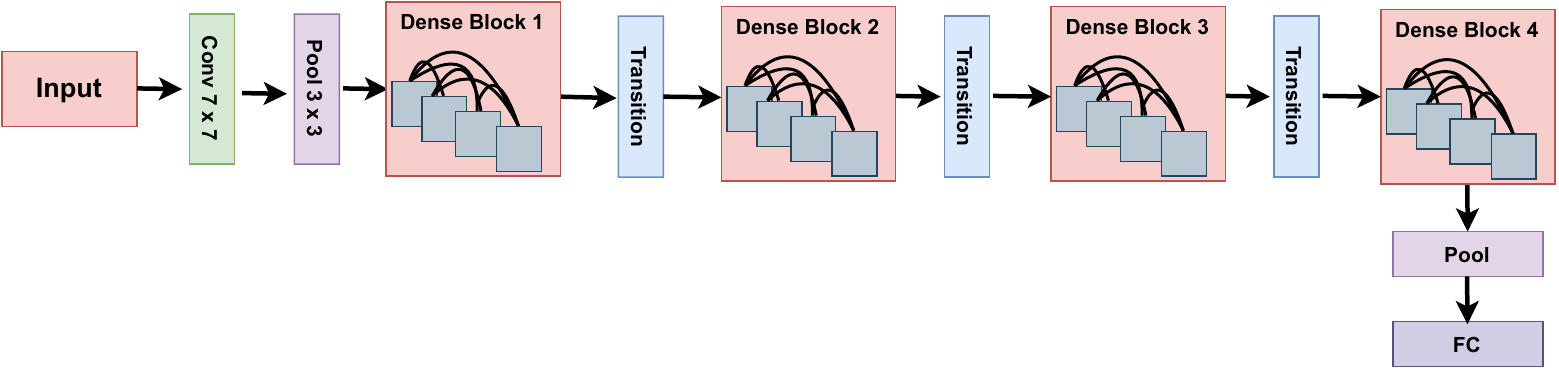}
	}}
	\caption{Internal architecture of DenseNet-121 classifier}
\vspace{3mm}	
\label{DenseNet}
\end{figure*}

DenseNet\cite{b9_huang} consists of convolutional layers connected to their previous layers in a feed-forward manner. The majority of CNN designs suffer from the vanishing gradient problem, which occurs when the Rectified Linear Unit (ReLU) is used as an activation function. In deeper CNNs, the gradient of the loss function approaches zero due to the use of ReLU activation functions. This phenomenon further impacts the learning process which results in overfitting of the model. As fingerprint images have limited color and texture information, typical CNNs encounter this issue when they are utilized in this domain. This problem of vanishing gradient is solved by DenseNet which has skip connections among the layers inside convolutional blocks as shown in Fig. \ref{DenseNet}.
%\vspace{4 mm}

\begin{figure*}[!bht]
	\centering
        \resizebox{0.75\textwidth}{!}{
	\subfigure[Dense Block \hspace{45 mm} (b) Transition Block ]
	{	
		\includegraphics[width=0.60\textwidth]{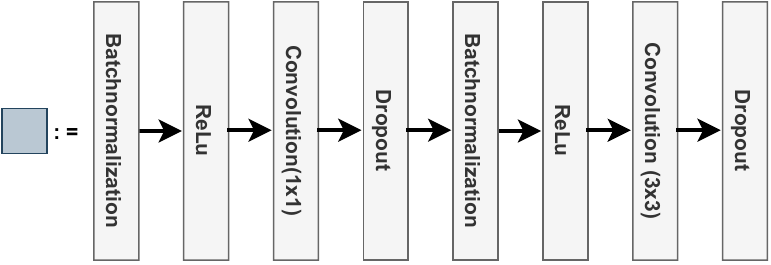}
		\hspace{15 mm}
		\includegraphics[width=0.42\textwidth ]{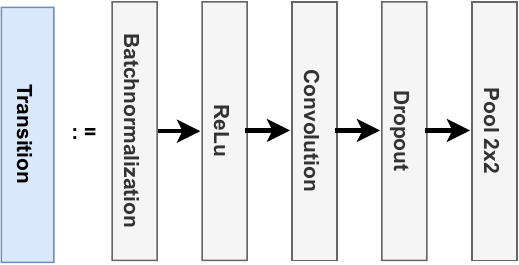}
	}}
	\caption{Composition of (a) dense block and (b) transition block}
	\label{Densenet_internal}
	\vspace{-2mm}
\end{figure*}

DenseNet-121 comprises four dense blocks containing six, twelve, twenty-four, and sixteen convolutional layers, respectively. The architecture of DenseNet is illustrated in Figure \ref{DenseNet}. As per this architecture, each dense block performs three operations: batch normalization, ReLU activation and convolution, and aggregates feature maps from all its preceding layers. The output of a dense block is formulated as shown in Eq. (\ref{eqn:densenet}).

\begin{equation}
\label{eqn:densenet}
X_n = F_n[A_0, A_1, A_2, A_3, .....A_{n_-1}] 
\end{equation}

where $F_{n}$ denotes a function that performs batch normalization followed by a convolution operation and $A_0, A_1, A_{n-1}$ are representing the concatenation of all feature maps from the previous layers. In addition, after each dense block, a transition block is utilized to execute a convolution operation with a kernel size of $1\times1$. Following this convolution procedure, a pooling operator minimizes the size of feature maps after each dense block. Figure \ref{Densenet_internal} depicts the internal architecture of the dense layer and transition layer.

\subsection{Deep Neural Network (DNN)}
DNN is an artificial neural network that has several fully connected layers. The working of DNNs is inspired by the human brain as its neurons resemble similar characteristics as neurons of the brain. We have used DNN to test the individual performance of the handcrafted features and LPQ image descriptor and to design the proposed architecture.

\subsection{Proposed DyFFPAD Model}

\begin{figure*}[!bht]
	\centering
	\resizebox{0.50\textheight}{!}{
		{
			
			\includegraphics[]{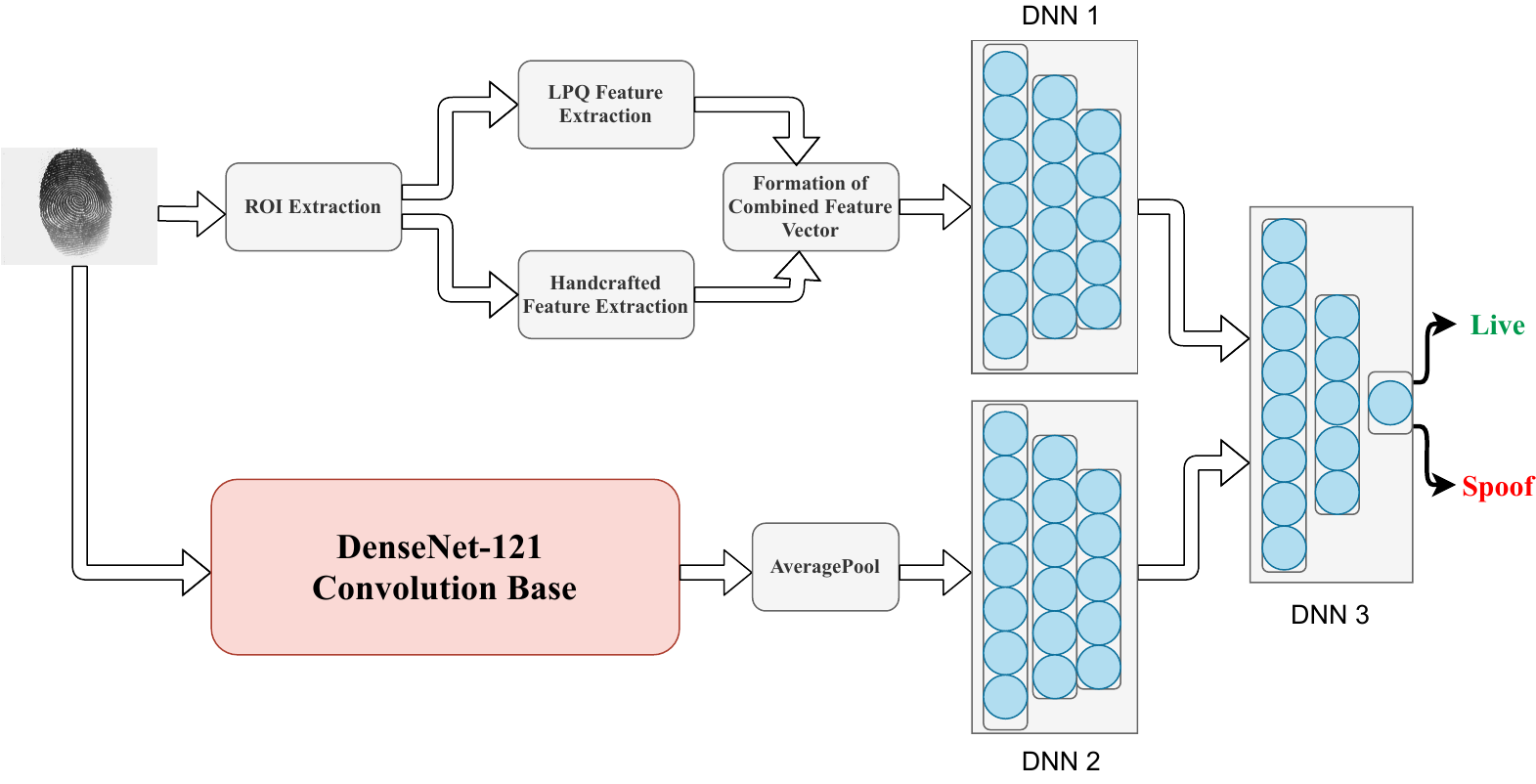}
	}}
	\caption{Block diagram of DyFFPAD architecture}
\vspace{3mm}	
\label{DyFFPAD_main}
\end{figure*}

The architecture of the proposed DyFFPAD is depicted in Fig.~\ref{DyFFPAD_main}. It is a dynamic ensemble of handcrafted and convolutional features. The DNN works on the feature vectors generated from handcrafted features and LPQ image descriptors while the convolution base of DenseNet extracts convolutional features from the input samples. The DNN 1 in Fig. \ref{DyFFPAD_main} which consists of three dense layers having 256, 128, and 32, neurons, respectively is fed with the features extracted from LPQ image descriptor and handcrafted features. Similarly, DNN 2 is composed of three dense layers having 1024, 256, and 32, neurons, respectively. Finally, the outcome of both DNNs is combined to form a feature vector which is further fed as an input to another DNN 3 that consists of fully connected layers with 48, 16, and 1 neurons, respectively. In the forward pass of the training process, a confidence score is produced by the model for an input sample. The loss of the output score with respect to the expected output is then calculated which is back-propagated to the entire model including DenseNet as well as DNNs for the learning of their parameters. The proposed model works in an end-to-end manner for the forward and backward pass of the training process and reduces the limitations of the static ensemble of handcrafted features and convolutional features for the accomplishment of a classification task.

\subsubsection{Pre-processing of input samples} The fingerprint samples captured with the sensing devices have white space around the fingertip impression. This white space is meaningless and is required to be removed for the extraction of the desired features from fingerprint samples. Therefore, a region of interest from the input fingerprint sample is identified as in pre-processing.
\subsubsection{Feature Extraction} The LPQ and handcrafted features are extracted from the pre-processed fingerprints to form a combined feature vector. The details of the utilized features are described in sections \ref{LPQ1} and \ref{handc}.

\subsubsection{Training of DyFFPAD}
The proposed DyFFPAD model is trained on benchmark LivDet databases from scratch. To expedite the training process, we initialized the parameters of DenseNet's convolution base with ImageNet weights instead of random values. The results of the proposed model under intra-sensor, same-material, and cross-material protocols are discussed in Section \ref{experimentalresults}.

\section{Experimental Setup}
\label{experimentalresults}
\subsection{\textbf{Database}}
The proposed work is evaluated on the LivDet 2015, 2017, and 2019 databases to assess the performance widely used publicly available database. Each database includes multiple sensing devices, with fingerprint samples from these devices organized into separate datasets for training and testing. The details of utilized databases including the name of sensing devices, number of samples in training and testing, and names of the fabrication materials are mentioned in Table \ref{Database_details}. \\
\begin{table*}[!ht]
\begin{center}
\caption{Details of the databases utilized for the validation of DyFFPAD}
\label{Database_details}
\resizebox{0.90\textwidth}{!}{
\begin{tabular}{p{3.5 cm}lccl}
\hline
\textbf{Database}                     & \textbf{Sensing device}          & \textbf{Spoof (Train/Test)} & \textbf{Live (Train/Test)} & \textbf{Spoofing Materials}                                                      \\ \hline

{\textbf{LivDet 2015 \cite{b34_livdet2015}}} & \textbf{Crossmatch}       & 1473/1448 & 1000/1500         &  Ecoflex, Playdoh, Body Double, OOMOO, Gelatine                                     \\ \cline{2-5} 
                                      
                                      & \textbf{Greenbit} & 1000/1500            & 1000/1000      &                                                                                    \\ \cline{2-4}& \textbf{Digital Persona} & 1000/1500   & 1000/1000        & \multirow{3}{*}{Gelatine, Woodglue, Ecoflex, Latex,  Liquid Ecoflex,   RTV}         \\ \cline{2-4}
                                      & \textbf{Biometrika} & 1000/1500             & 1000/1000      &                                                                                    \\ \hline
{\textbf{LivDet 2017 \cite{livdet_2017}}} & \textbf{Greenbit} & 1200/2040           & 1000/1700       & \multirow{3}{*}{Gelatine, Latex,   Liquid Ecoflex, Body Double, Ecoflex, Woodglue } \\ \cline{2-4}
                                      & \textbf{Orcanthus}  & 1180/2018          & 1000/1700      &                                                                                    \\ \cline{2-4}
                                      & \textbf{Digital Persona}& 1199/2028    & 999/1692         &                                                                                    \\ \hline
{\textbf{LivDet 2019 \cite{b30_orru}}} & \textbf{Greenbit}  & 1200/1224           & 1000/1020      &   Woodglue, Mix1, Mix2, Liquid Ecoflex ,         Body Double, Ecoflex                                                                       \\ \cline{2-5} 
                                      & \textbf{Orcanthus}  & 1200/1088        & 1000/990         & Mix1, Mix2, Liquid Ecoflex,  Body Double, Ecoflex, Woodglue                                                                                 \\ \cline{2-5} 
                                      & \textbf{Digital Persona} & 1000/1224     & 1000/1099      & Mix1, Mix2, Liquid Ecoflex,    Ecoflex, Gelatine, Woodglue, Latex                                                                               \\ \hline
\end{tabular}}
\end{center}
\end{table*}
\vspace{-3mm}

\subsection{Performance Metrics}

The metrics specified by ISO/IEC IS 30107 \cite{misc1} are used to assess the performance of the proposed model. The performance is measures using Attack Presentation Classification Error Rate (APCER) and Bonafide Presentation Classification Error Rate (BPCER) metrics. APCER measures the proportion of spoof fingerprint samples that are misclassified. The proportion of real fingerprint samples that are misclassified as spoofs is measured by BPCER.
Eq. (\ref{APCER}) and Eq. (\ref{BPCER}) denote APCER and BPCER, respectively.

\begin{equation}
\label{APCER}
APCER = \frac{\textnormal{\# Mis-classified spoof samples}}{\textnormal{\# fake samples}} \times{100}
\end{equation}

\begin{equation}
\label{BPCER}
BPCER = \frac{\textnormal{\# Mis-classified live samples}}{\textnormal{\# live samples}} \times {100}
\end{equation}
The overall performance is evaluated using Average Classification Error (ACE) which is determined by averaging APCER and BPCER. The formulation of ACE is represented by Eq. (\ref{ACE}).
\begin{equation}
\label{ACE}
ACE = \frac{APCER + BPCER}{2}
\end{equation}

The ACE is also used to determine the proposed model's accuracy which is formulated as Eq. (\ref{Accuracy}).
\begin{equation}
\label{Accuracy}
Accuracy = 100 -  ACE
\end{equation}
We have also reported the Equal Error Rate (EER) to measure the integrity of our proposed FPAD model. EER is observed by finding an appropriate threshold value where both the APCER and BPCER values are equal.

\subsection{\textbf{Implementation Details}}
The proposed model is implemented using the TensorFlow-Keras library in Python. The NVIDIA TESLA P100 GPU was used for all training and testing. Each model is trained for 300 epochs which requires 4 to 5 hours for its training. The learning rate and batch size are kept as 0.0001 and 8, respectively.

\subsection{\textbf{Ablation Study}} 
\begin{figure*}[!ht]
	\centering
	
	\resizebox{0.70\textwidth}{!}{
	\subfigure
	{
		\includegraphics[width=0.75\textwidth]{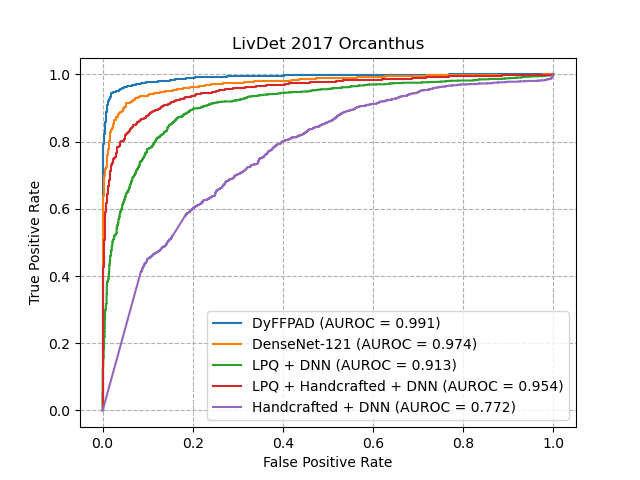}
		\hspace{3 mm}
		\includegraphics[width=0.75\textwidth]{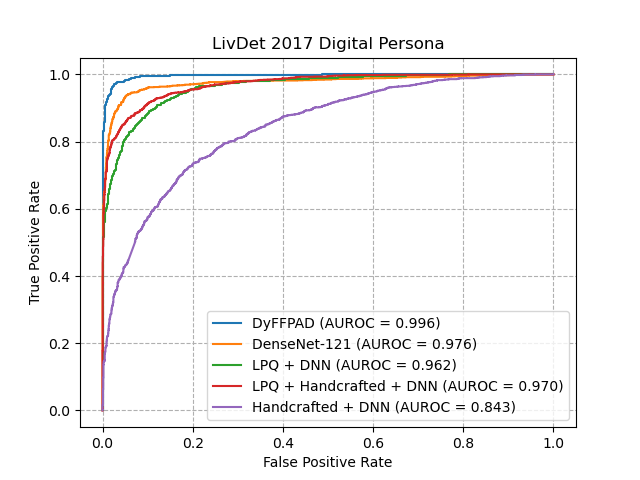}
	}}
	\resizebox{0.35\textwidth}{!}{
	\subfigure
	{	
		\includegraphics[width=0.75\textwidth]{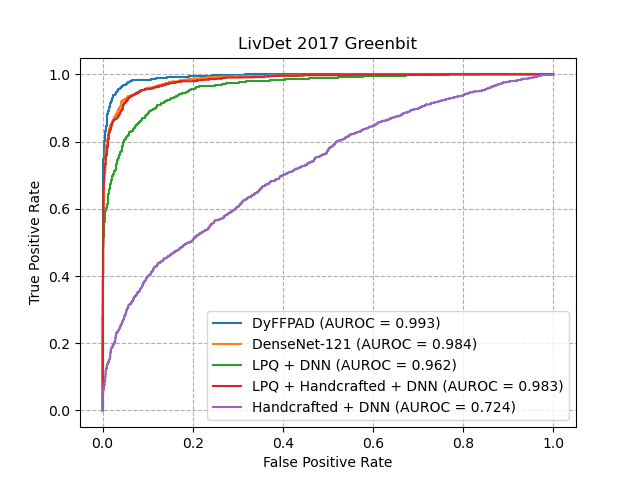}
		\hspace{3 mm}
	}}
	%\vspace{-4 mm}
	\caption{Receiver Operating Characteristic (ROC) curves for LivDet 2017 datasets (Orcanthus, Digital Persona, and Greenbit)}
	\label{ROC_Cur}
	\vspace{-2mm}
\end{figure*}

To assess, the performance of the proposed model with handcrafted features, LPQ image descriptor, and DenseNet classifier, we have trained the corresponding DNN and CNN models. The handcrafted features and LPQ obtained from the fingerprint samples are used for the training of DNN while DenseNet-121 is trained on original fingerprint samples. The performance of the individual models and the proposed model is validated using LivDet 2017 which is a challenging database since it contains spoof samples of different materials in training and testing sets. The results of this experiment are presented in Table \ref{tab: Ablation} which clearly indicates the supremacy of DyFFPAD over the DNN trained with LPQ, DNN trained with handcrafted features and LPQ, and the DenseNet classifier. The dynamic ensemble of handcrafted and deep features empowers the model to attain better classification accuracy irrespective of the materials used for the fabrication of spoofs. The proposed DyFFPAD models attain an overall classification accuracy of 96.55\% as compared with 79.12\% (LPQ+DNN), 75.93\% (Handcrafted+DNN), 90.91\% (LPQ+Handcrafted+DNN) and 93.64\% (DenseNet-121 classifier). The performance comparison of the aforementioned models with DyFFPAD is also shown using the Receiver Operating Characteristics (ROC) curve which is depicted in Fig. \ref{ROC_Cur}. ROC curve represents relationship between True Positive Rate (TPR) and False Positive Rate (FPR) and is essential in measuring the performance of a machine/deep learning model. The Area Under the Curve (AUC) represents the region enclosed by the plot of the FPR versus TPR. A larger AUC indicates the better proficiency of the model while classifying the input samples.

\section{\textbf{Experimental Results}}
\label{EXPCA}
The performance has been validated in two benchmark scenarios: intra-sensor (known spoofing material), and intra-sensor (unknown spoofing material). The following subsections provide detailed descriptions and results of these evaluation protocols.

\begin{table}[!bht]
\begin{center}
\caption{Findings of the ablation study performed on LivDet 2017 database in terms of accuracy}
\label{tab: Ablation}
\resizebox{0.48\textwidth}{!}{
\begin{tabular}{p{3.5 cm} c c c c}
\hline
\textbf{Method} & \textbf{\begin{tabular}[c]{@{}c@{}}Orcanthus \\ (\%)\end{tabular}}& \textbf{\begin{tabular}[c]{@{}c@{}}Digital Persona \\(\%)\end{tabular}} & \textbf{\begin{tabular}[c]{@{}c@{}}Greenbit \\(\%)\end{tabular}} & \multicolumn{1}{l}{\textbf{\begin{tabular}[c]{@{}c@{}}Average \\(\%) \end{tabular}}} \\ \hline

\textbf{LPQ+DNN}              &  84.91                                            & 77.06                                                  & 75.40                                            &    79.12   \\ \hline
\textbf{Handcrafted+DNN}              &     75                                        &      76.62                                            &     76.19                                        &     75.93                                                        \\ \hline
\textbf{Handcrafted+LPQ+DNN}              & 88.89                                            & 90.23                                                  & 93.62                                            & 90.91                                 \\ \hline
\textbf{DenseNet-121 Classifier}             & 92.79                                            & 94.19                                                 & 93.95                                            &  93.64                                \\ \hline
                 \textbf{DyFFPAD}           & \textbf{\textbf{96.34}}                                   & \textbf{\textbf{97.14}}                                         & \textbf{\textbf{96.00}}                                   & \textbf{\textbf{96.49}}                        \\ \hline
                 
\end{tabular}
}
\end{center}
\end{table}

\begin{table*}[!bht]
\caption{Experimental results of our proposed method on LivDet 2015 database}
\label{tab : 2015_Intra_Sensor}
\begin{center}
\resizebox{0.85\textwidth}{!}{
\begin{tabular}{llccccc}
\hline
\textbf{Database} & \textbf{Sensor} & \textbf{\begin{tabular}[c]{@{}c@{}}BPCER \\ (\%)\end{tabular}} & \textbf{\begin{tabular}[c]{@{}l@{}}APCER\\ (Known) (\%)\end{tabular}} & \textbf{\begin{tabular}[c]{@{}l@{}}APCER \\ (Unknown) (\%)\end{tabular}} & \textbf{ACE (\%)} & \textbf{EER (\%)} \\ \hline
\multirow{5}{*}{\textbf{LivDet 2015}} & \textbf{Digital Persona} & 8.1 & 4.25& 5.74 &6.08 & 6.2 \\ \cline{2-7} 
 &\textbf{Crossmatch} & 0.83 &  0.50 & 0.68 & 0.72 & 0.8\\ \cline{2-7} 
 & \textbf{Biometrika} & 4.5 & 2.5 & 3.62 &3.52 & 3.6 \\ \cline{2-7} 
 & \textbf{Greenbit} & 9.75 &1.9  &3.1 & 5.3 & 7.9 \\ \cline{2-7} 
 & \textbf{Average} & \textbf{5.79} & \textbf{2.28} & \textbf{3.28} & \textbf{3.90} & \textbf{4.57} \\ \hline
\end{tabular}}
\end{center}
\end{table*}

\subsubsection{\textbf{Intra-Sensor (Known Spoofing Material)}} 
The samples of the training and testing sets are acquired using the same sensing device and they are fabricated using the same materials in both the sets. LivDet 2015 partially follows this protocol as it consists of two-thirds of the testing spoofs fabricated with known materials and one-third with unknown materials. The results on the LivDet 2015 database are presented in Table \ref{tab : 2015_Intra_Sensor}. The results presented in table shows that the proposed model achieves an average APCER (known) of 2.28\% and BPCER of 5.79\%. 

\begin{table*}[!bht]
\caption{Experimental results of our proposed method on LivDet 2017 and 2019 databases}
\label{tab: intra_2017_2019}
\begin{center}
\resizebox{0.70\textwidth}{!}{
\begin{tabular}{llcccc}
\hline
\textbf{Database} & \textbf{Sensor} & \textbf{\begin{tabular}[c]{@{}c@{}}BPCER\\ (\%)\end{tabular}} & \textbf{\begin{tabular}[c]{@{}c@{}}APCER \\ (\%)\end{tabular}} & \textbf{\begin{tabular}[c]{@{}c@{}}ACE \\(\%) \end{tabular}} & \textbf{\begin{tabular}[c]{@{}c@{}}EER \\(\%)\end{tabular}} \\ \hline
\multirow{4}{*}{\textbf{LivDet 2017}}  
 & \textbf{Orcanthus} &5.59  & 2.04 &3.66  & 4.2\\ \cline{2-6} 
 & \textbf{Digital Persona} &3.34  & 2.46 &2.87  & 2.7  \\ \cline{2-6}
 & \textbf{Greenbit} & 4.68 & 3.44 &  4.00  & 3.9\\ \cline{2-6} 
 & \textbf{Average} & \textbf{4.53} & \textbf{2.64} & \textbf{3.51} & \textbf{ 3.6} \\ \hline
\multirow{4}{*}{\textbf{LivDet 2019}} & \textbf{Digital Persona} & 12.86 &4.74  & 8.43  & 9.5\\ \cline{2-6} 
 & \textbf{Greenbit} & 2.06 & 1.80 & 1.92 & 1.8\\ \cline{2-6} 
 & \textbf{Orcanthus} & 2.83 & 0.65 &  1.69 & 1.8\\ \cline{2-6} 
 & \textbf{Average} & \textbf{5.91} & \textbf{2.39} & \textbf{4.01}  &\textbf{4.23}\\ \hline
\end{tabular}
}
\end{center}
\end{table*}

\subsubsection{\textbf{Intra-Sensor (Unknown Spoofing Material)}}
In this experimental protocol, the samples of the training and testing fingerprint samples are acquired using the same sensing device but the fabrication material used is different in both sets. This validation protocol is designed to evaluate the FPAD system's efficacy in a real-attack scenario, wherein an adversary may attempt to present a forged fingerprint artifact created with novel fabrication materials that were not part of the FPAD model's training data. LivDet 2017 and 2019 are prepared as per this protocol of FPAD. 
The results of this evaluation protocol are presented in Table \ref{tab: intra_2017_2019}. As given in Table \ref{tab: intra_2017_2019}, our model achieves a BPCER of 4.53\%, APCER of 2.64\%, and an ACE of 3.5\%, and EER of 3.6\% on the LivDet 2017 database. Similarly, on the LivDet 2019 database, the proposed model showcases the ability to differentiate between live and spoof samples, with a BPCER of 5.91\%, APCER of 2.39\%, and EER of 4.23\%. Notably, when confronted with the spoof samples contained in the LivDet 2015 database, the proposed method exhibits an average APCER (unknown) of 3.28\%, as indicated in Table $\ref{tab : 2015_Intra_Sensor}$. It also shows that the proposed methods achieve an ACE of 3.90\% and EER of 4.57\% on LivDet 2015 database.
The live and spoof fingerprint samples exhibit different textures and ridge valley widths due to significant differences between the properties of finger skin and spoofing materials. The convolutional layers of CNNs allow for effective classification by extracting distinguishing features from spoof samples, regardless of the materials used for their fabrication. Moreover, the handcrafted features and image descriptor extract discriminating features for the training of DNN. The dynamic ensemble of DNN and CNN facilitates improved parameter learning for both networks. The proposed architecture reduces the need for training two different models. A comparative study of the obtained results of our proposed method on benchmark databases is provided in section \ref{comparative analysis}.

\subsection{\textbf{Comparative Analysis \label{comparative analysis}}} In this section, state-of-the-art methods are compared with the results of the proposed method in different evaluation conditions. A detailed comparative study is provided in the following subsections.

\begin{table}[!hbt]
\begin{center}
\caption{Comparison with state-of-the-art methods on LivDet 2015}
\label{tab: Comparison_Intra_2015} 
\resizebox{0.49\textwidth}{!}{
\begin{tabular}{p{2.2cm}ccccc}
\hline
 \textbf{Method}                    & \textbf{\begin{tabular}[c]{@{}c@{}}Crossmatch \\ (\%)\end{tabular}} & \textbf{\begin{tabular}[c]{@{}c@{}}Greenbit \\ (\%)\end{tabular}} & \textbf{\begin{tabular}[c]{@{}c@{}}Digital Persona \\ (\%)\end{tabular}} & \textbf{\begin{tabular}[c]{@{}c@{}}Biometrika \\ (\%)\end{tabular}} & \textbf{\begin{tabular}[c]{@{}c@{}}Average \\(\%) \end{tabular}}                                          
 \\ \hline
  \cite{b4_uliyan}                                                                                     & 95.00                                                                    & -                                                                       & -                                                                            & -                                                                          & 95.00                        \\ \hline
                                    
 \cite{b5_sharma1}                                                                                    & 98.07                                                                    & 95.7                                                                    & 94.16                                                                        & 95.22                                                                      & 95.78                        \\ \hline
 \cite{b47_jung2}                                                                                       & 98.60                                                                    & 96.20                                                                   & 90.50                                                                        & 95.80                                                                      & 95.27                        \\ \hline
 
 \cite{b37_deepika}                                                                                    & 95.0                                                                    & 96.7                                                                    & 97.2                                                                        & 95.50                                                                      & 96.10
\\ \hline
\cite{xia_2}                                                                                     &      89.18                                                               &    95.47                                                                &      86.28                                                                    &              90.36                                                          &   90.32
                \\  \hline

LivDet 2015 Winner \cite{b34_livdet2015}                                                                          & 98.10                                                                    & 95.40                                                                   & 93.72                                                                        & 94.36                                                                      & 95.39                        \\ \hline

            \cite{Yuan2020}                                                                                     &   96.54                                                                  &    95.32                                                                &  93.2                                                                        &  93.76                                                                      &  94.70 
                \\ \hline
                 \cite{b31_zhang}                                                                                      & 97.01                                                                    & 97.81                                                                   & 95.42                                                                        & 97.02                                                                      & 96.82 
                 \\ \hline
                 \cite{anuj_1}                                                                                      &     97.51                                                                &     97.63                                                               &     93.50                                                                    &    96.18                                                                   &  96.20
                \\ \hline

                 \cite{Rubab}                                                                                      &   97.9                                                                  &  97.48                                                                  &    88.96                                                                     &  97.9                                                                     & 95.5 \\ \hline

             \cite{anuj_5}                                                                                      &     95.01                                                                &    96.95                                                                &   93.32                                                                      &    96.00                                                                   & 95.32 \\ \hline

                                                                                                \cite{BineetKaur}    & -                                                                    & -                                                                   &  -                                                                       &                                                                       &96.49  \\ \hline
             { \textbf{DyFFPAD}}                                                              & { \textbf{99.29}}                                    & { \textbf{94.71}}                                   & { \textbf{93.92}}                                        & { \textbf{96.48}}                                      & {\textbf{96.10}} \\ \hline
\end{tabular}}
\end{center}
\end{table}

\subsubsection{\textbf{Comparative study on LivDet 2015 database}}
 Table \ref{tab: Comparison_Intra_2015} provides a detailed comparison of the proposed method's performance with state-of-the-art methods. The results presented in Table \ref{tab: Comparison_Intra_2015} clearly show that the proposed method achieves an accuracy of 99.29\% which is the best as compared with all other methods validated on crossmatch sensor dataset while it performs better than all other methods except \cite{Rubab} and \cite{b31_zhang} while being validated on biometrika dataset. In terms of overall performance, The proposed method outperforms the methods presented in \cite{b5_sharma1}, \cite{b45_jung}, \cite{b34_livdet2015}, \cite{b4_uliyan}, \cite{b37_deepika}, \cite{xia_2}, \cite{Yuan2022}, and \cite{anuj_5} while it is comparable to the methods discussed in \cite{BineetKaur}, \cite{anuj_1}, and \cite{b31_zhang}. 

\subsubsection{\textbf{Comparative study on LivDet 2017 database}}
The performance of the proposed method is also compared with methods validated on the LivDet 2017 database. Our method successfully demonstrates robust classification capabilities in distinguishing between live and spoof samples, achieving commendable accuracy. The comparison shown in Table \ref{tab: Comparison_Intra_2017} indicate that our method is able to perform better as compared with the methods discussed in \cite{b14_chugh1}, \cite{b15_chugh2}, \cite{b31_zhang}, \cite{anuj_4}, and \cite{GONZÁLEZ-SOLER} for orcanthus and digital persona sensors while it is comparable with \cite{anuj_1} on orcanthus sensor. The proposed method achieves an accuracy of 96.00\% on greenbit dataset which is better than of \cite{GONZÁLEZ-SOLER}, \cite{b31_zhang}, \cite{anuj_4} and is comparable with \cite{b15_chugh2}. The overall classification accuracy achieved by the proposed method is 96.49\% which is better than all other methods. This comparison reveals that the dynamic ensemble of handcrafted and convolutional features is able to detect spoofs irrespective of the material used for their fabrication.

\begin{table}[!bht]
\begin{center}
\caption{Comparison with state-of-the-art methods on LivDet 2017 database}
\label{tab: Comparison_Intra_2017}
\resizebox{0.49\textwidth}{!}{
\begin{tabular}{p{2.5 cm} c c c c}
\hline
\textbf{Method} & \textbf{\begin{tabular}[c]{@{}c@{}}Orcanthus \\ (\%)\end{tabular}}& \textbf{\begin{tabular}[c]{@{}c@{}}Digital Persona \\ (\%)\end{tabular}} & \textbf{\begin{tabular}[c]{@{}c@{}}Greenbit \\ (\%)\end{tabular}} & \multicolumn{1}{l}{\textbf{\begin{tabular}[c]{@{}c@{}}Average \\ (\%) \end{tabular}}} \\ \hline
 \cite{b31_zhang}              & 93.93                                            & 92.89                                                  & 95.20                                            & 94.00                                 \\ \hline

 \cite{b14_chugh1}              & 95.01                                           &   95.20                                                 & 97.42                                            & 95.88                                 \\ \hline
 \cite{GONZÁLEZ-SOLER}              & 94.38                                            & 95.08                                                  & 94.54                                            & 94.66                                 \\ \hline

 \cite{b15_chugh2}              & 94.51                                            & 95.12                                                  & 96.68                                            & 95.43                                 \\ \hline
\cite{anuj_1}                                                                                      &    96.53                                                                 &   93.64                                                                 & 94.74                                                                        &       94.97                                                                  
                \\ \hline
    
                \cite{anuj_4}                                                                                      &  95.94                                                                   &    95.40                                                                &   93.79                                                                      &  95.05                                                                       \\  \hline

                 \textbf{DyFFPAD}           & \textbf{96.34}                                   & \textbf{97.14}                                         & \textbf{96.00}                                   & \textbf{96.49}                        \\ \hline
\end{tabular}
}
\end{center}
\end{table}

\subsubsection{\textbf{Comparative study on LivDet 2019 database}}
Table \ref{tab: Comparison_Intra_2019} compares the performance of the proposed model on the LivDet 2019 database. This comparison reveals that the proposed method achieves an accuracy of 91.57\%, outperforming the approaches outlined in \cite{b15_chugh2} as well as the FPAD algorithms, including JWL LivDet, JungCNN, and ZJUT DET, when tested on samples captured with digital persona sensor. For the orcanthus sensor, the proposed method outperforms all other methods with a classification accuracy of 98.32\% while it remains second after \cite{GONZÁLEZ-SOLER} with an accuracy of 91.57\% on digital persona dataset. The overall performance comparison indicates that the proposed method outperforms all other methods except, \cite{GONZÁLEZ-SOLER}, achieving a classification accuracy of 95.99%.

The comparative analysis mentioned above concludes that the proposed method consistently outperforms state-of-the-art methods in the intra-sensor paradigm of FPAD, for both known or unknown materials scenario. As compared to standard CNN-based techniques, the dynamic ensemble of DNN empowered with handcrafted features and CNN enables them to learn their parameters better.

\begin{figure*}[!bht]
	\centering
	
	\resizebox{0.70\textwidth}{!}{
	\subfigure
	{
		\includegraphics[width=0.75\textwidth]{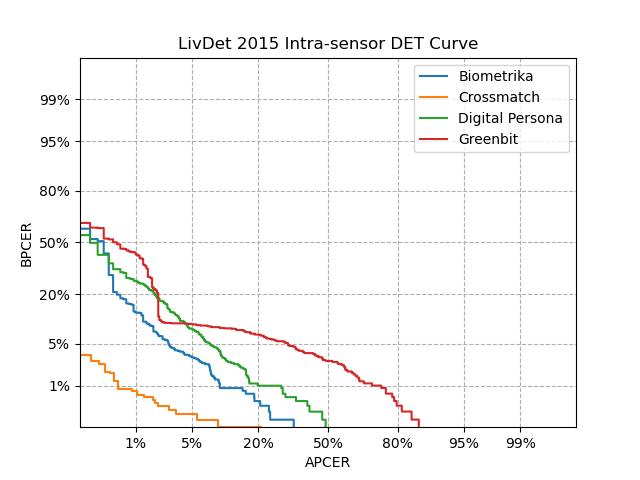}
		\hspace{4 mm}
		\includegraphics[width=0.75\textwidth]{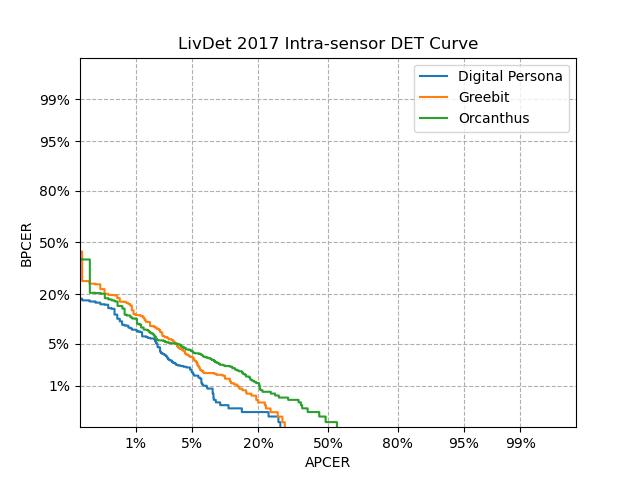}
	}}
	\resizebox{0.35\textwidth}{!}{
	\subfigure
	{	
		\includegraphics[width=0.75\textwidth]{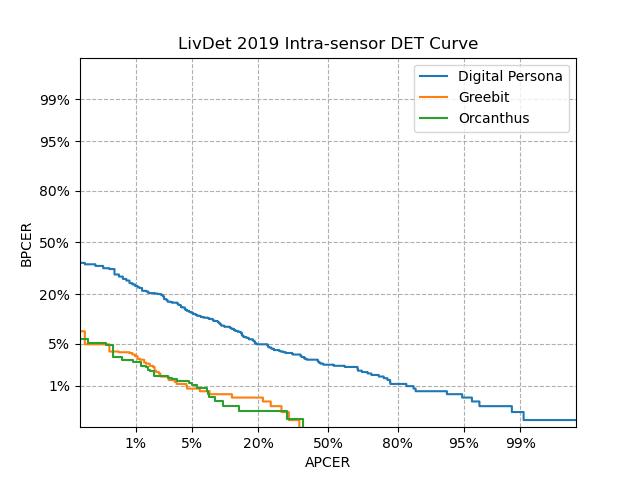}
		\hspace{4 mm}
	}}
	%\vspace{-4 mm}
	\caption{Detection Error Trade-off (DET) curves for LivDet 2015, 2017, and 2019 databases}
	\label{DET_Cur}
	\vspace{-2mm}
	
\end{figure*}

\begin{table}[!bht]
\begin{center}
\caption{Comparison with state-of-the-art methods on LivDet 2019 database}
\label{tab: Comparison_Intra_2019}
\resizebox{0.49\textwidth}{!}{
\begin{tabular}{p{2.5 cm} c c c c}
\hline
\textbf{Method} & \textbf{\begin{tabular}[c]{@{}c@{}}Orcanthus \\ (\%)\end{tabular}}& \textbf{\begin{tabular}[c]{@{}c@{}}Digital Persona \\ (\%)\end{tabular}} & \textbf{\begin{tabular}[c]{@{}c@{}}Greenbit \\ (\%)\end{tabular}} & \multicolumn{1}{l}{\textbf{Average (\%)}} \\ \hline

\cite{b15_chugh2}              &    97.50                                        &     83.64                                              &99.73                                             & 93.62                                \\ \hline

JWL LivDet \cite{b30_orru}              &      97.45                                      &       88.86                                            & 99.20                                            &   95.17                              \\ \hline
ZJUT Det A \cite{b30_orru}              &       97.50                                     &      88.77                                             &    99.20                                         &              95.16                   \\ \hline

 \cite{anuj_1}                                                                                      &    97.55                                                                 & 81.55                                                                   &    98.71                                                                     &        92.9                                                                
                \\ \hline
    
                \cite{anuj_4}                                                                                      &   97.72                                                                  &     90.59                                                               &      97.31                                                                   &    95.20                                                                    \\ \hline

                 \cite{GONZÁLEZ-SOLER}                                                                                      &     97.21                                                                &   97.68                                                                 &      93.63                                                                   &   96.17                                                                      \\ \hline

                 \textbf{DyFFPAD}           & \textbf{98.32}                                   & \textbf{91.57}                                         & \textbf{98.08}                                   & \textbf{95.99}          \\ \hline
\end{tabular}}
\end{center}
\end{table}

\subsection{Evaluation of DyFFPAD in High-Security Systems}

In the context of high-security systems, it is imperative to evaluate the performance of an FPAD model comprehensively. While the model's fundamental objective is indeed to minimize the APCER, BPCER, and ACE, its assessment should extend beyond these metrics to ensure its efficacy in safeguarding high-security environments. In this research, we report the findings of the suggested model using the Detection Error Trade-off (DET) curve, which is a graphical depiction of the error rates produced by a binary classification system when the classification threshold value is varied.
The DET curves for all datasets from the LivDet 2015, 2017, and 2019 databases are shown in Fig. \ref{DET_Cur}. As shown in  Fig. \ref{DET_Cur} it is evident that the proposed model achieves a BPCER of less than 1\% to achieve APCER of 1\% when evaluated on crossmatch, and it is in the range of 15\% - 40\% for biometrika, greenbit and digital persona sensors of LivDet 2015 database. On the LivDet 2017 database, the model is able to restrict BPCER in the range of 7\% - 18\% when testing spoof samples are obtained using unknown spoof materials. Similarly, the model maintains a BPCER of less than 5\% for greenbit and orcanthus sensors of LivDet 2019 database.

\section{Conclusion \label{conclusion}}
In this paper, a novel end-to-end method is presented which utilizes the handcrafted and convolutional features together for the detection of PAs. The increasing deployment of AFRS in security and commercial applications makes them vulnerable to various possible PAs. The proposed method has demonstrated the capability to detect PAs fabricated with known and unknown fabrication materials. Additionally, it suggests a novel ensemble of deep and handcrafted features in an end-to-end manner, making it appropriate for real-time collaboration with the AFRS. The proposed method is tested on fingerprint databases which are prepared as per benchmark experimental protocols. In the future, we will investigate the proposed model's capabilities for cross-sensor validation using benchmark fingerprint databases.

\section*{Declaration}
\begin{itemize}
    \item The publicly accessible datasets used to validate this work can be obtained at [https://livdet.org/registration.php].
\end{itemize}

\begin{itemize}
    \item Conflict of Interest: The authors declare that there are no conflicts of interest.
\end{itemize}

\printcredits

%% Loading bibliography style file
%\bibliographystyle{model1-num-names}
%\bibliographystyle{cas-model2-names}
\bibliographystyle{cas-model2-names}

% Loading bibliography database
\bibliography{cas-refs}

\begin{thebibliography}{51}
\expandafter\ifx\csname natexlab\endcsname\relax\def\natexlab#1{#1}\fi
\providecommand{\url}[1]{\texttt{#1}}
\providecommand{\href}[2]{#2}
\providecommand{\path}[1]{#1}
\providecommand{\DOIprefix}{doi:}
\providecommand{\ArXivprefix}{arXiv:}
\providecommand{\URLprefix}{URL: }
\providecommand{\Pubmedprefix}{pmid:}
\providecommand{\doi}[1]{\href{http://dx.doi.org/#1}{\path{#1}}}
\providecommand{\Pubmed}[1]{\href{pmid:#1}{\path{#1}}}
\providecommand{\bibinfo}[2]{#2}
\ifx\xfnm\relax \def\xfnm[#1]{\unskip,\space#1}\fi
%Type = Misc
\bibitem[{30107-3:2017(en)(2017)}]{misc1}
\bibinfo{author}{30107-3:2017(en), I.}, \bibinfo{year}{2017}.
\newblock \bibinfo{title}{Information technology — biometric presentation attack detection — part 3: Testing and reporting}.
%Type = Article
\bibitem[{Abhyankar and Schuckers(2009)}]{b19_abhyankar}
\bibinfo{author}{Abhyankar, A.}, \bibinfo{author}{Schuckers, S.}, \bibinfo{year}{2009}.
\newblock \bibinfo{title}{Integrating a wavelet based perspiration liveness check with fingerprint recognition}.
\newblock \bibinfo{journal}{Pattern Recognition} \bibinfo{volume}{42}, \bibinfo{pages}{452--464}.
%Type = Inproceedings
\bibitem[{Baishya et~al.(2023)Baishya, Tiwari, Rai and Dey}]{Anuj_2}
\bibinfo{author}{Baishya, J.}, \bibinfo{author}{Tiwari, P.K.}, \bibinfo{author}{Rai, A.}, \bibinfo{author}{Dey, S.}, \bibinfo{year}{2023}.
\newblock \bibinfo{title}{Impact of existing deep cnn and image descriptors empowered svm models on fingerprint presentation attacks detection}, in: \bibinfo{booktitle}{Proceedings of International Conference on Frontiers in Computing and Systems}, \bibinfo{publisher}{Springer Nature Singapore}, \bibinfo{address}{Singapore}. pp. \bibinfo{pages}{241--251}.
%Type = Article
\bibitem[{Chugh et~al.(2018)Chugh, Cao and Jain}]{b15_chugh2}
\bibinfo{author}{Chugh, T.}, \bibinfo{author}{Cao, K.}, \bibinfo{author}{Jain, A.K.}, \bibinfo{year}{2018}.
\newblock \bibinfo{title}{Fingerprint spoof buster: Use of minutiae-centered patches}.
\newblock \bibinfo{journal}{IEEE Transactions on Information Forensics and Security} \bibinfo{volume}{13}, \bibinfo{pages}{2190--2202}.
%Type = Article
\bibitem[{Chugh and Jain(2021)}]{b14_chugh1}
\bibinfo{author}{Chugh, T.}, \bibinfo{author}{Jain, A.K.}, \bibinfo{year}{2021}.
\newblock \bibinfo{title}{Fingerprint spoof detector generalization}.
\newblock \bibinfo{journal}{IEEE Transactions on Information Forensics and Security} \bibinfo{volume}{16}, \bibinfo{pages}{42--55}.
%Type = Inproceedings
\bibitem[{Deng et~al.(2009)Deng, Dong, Socher, Li, Li and Fei-Fei}]{imagenet}
\bibinfo{author}{Deng, J.}, \bibinfo{author}{Dong, W.}, \bibinfo{author}{Socher, R.}, \bibinfo{author}{Li, L.J.}, \bibinfo{author}{Li, K.}, \bibinfo{author}{Fei-Fei, L.}, \bibinfo{year}{2009}.
\newblock \bibinfo{title}{Imagenet: A large-scale hierarchical image database}, in: \bibinfo{booktitle}{2009 IEEE Conference on Computer Vision and Pattern Recognition}, pp. \bibinfo{pages}{248--255}.
%Type = Article
\bibitem[{Deng(2012)}]{mnist}
\bibinfo{author}{Deng, L.}, \bibinfo{year}{2012}.
\newblock \bibinfo{title}{The mnist database of handwritten digit images for machine learning research [best of the web]}.
\newblock \bibinfo{journal}{IEEE Signal Processing Magazine} \bibinfo{volume}{29}, \bibinfo{pages}{141--142}.
%Type = Article
\bibitem[{Derakhshani et~al.(2003)Derakhshani, Schuckers, Hornak and O'Gorman}]{Derakshini2003}
\bibinfo{author}{Derakhshani, R.}, \bibinfo{author}{Schuckers, S.A.}, \bibinfo{author}{Hornak, L.A.}, \bibinfo{author}{O'Gorman, L.}, \bibinfo{year}{2003}.
\newblock \bibinfo{title}{Determination of vitality from a non-invasive biomedical measurement for use in fingerprint scanners}.
\newblock \bibinfo{journal}{Pattern Recognition} \bibinfo{volume}{36}, \bibinfo{pages}{383--396}.
%Type = Inproceedings
\bibitem[{Espinoza and Champod(2011)}]{b16_espinoza}
\bibinfo{author}{Espinoza, M.}, \bibinfo{author}{Champod, C.}, \bibinfo{year}{2011}.
\newblock \bibinfo{title}{Using the number of pores on fingerprint images to detect spoofing attacks}, in: \bibinfo{booktitle}{2011 International Conference on Hand-Based Biometrics}, pp. \bibinfo{pages}{1--5}.
%Type = Inproceedings
\bibitem[{Frassetto~Nogueira et~al.(2014)Frassetto~Nogueira, de~Alencar~Lotufo and Campos~Machado}]{Nogueira2014}
\bibinfo{author}{Frassetto~Nogueira, R.}, \bibinfo{author}{de~Alencar~Lotufo, R.}, \bibinfo{author}{Campos~Machado, R.}, \bibinfo{year}{2014}.
\newblock \bibinfo{title}{Evaluating software-based fingerprint liveness detection using convolutional networks and local binary patterns}, in: \bibinfo{booktitle}{2014 IEEE Workshop on Biometric Measurements and Systems for Security and Medical Applications (BIOMS) Proceedings}, pp. \bibinfo{pages}{22--29}.
%Type = Inproceedings
\bibitem[{Ghiani et~al.(2012)Ghiani, Marcialis and Roli}]{LPQ}
\bibinfo{author}{Ghiani, L.}, \bibinfo{author}{Marcialis, G.}, \bibinfo{author}{Roli, F.}, \bibinfo{year}{2012}.
\newblock \bibinfo{title}{Fingerprint liveness detection by local phase quantization}, pp. \bibinfo{pages}{537--540}.
%Type = Article
\bibitem[{González-Soler et~al.(2021)González-Soler, Gomez-Barrero, Chang, Pérez-Suárez and Busch}]{GONZÁLEZ-SOLER}
\bibinfo{author}{González-Soler, L.J.}, \bibinfo{author}{Gomez-Barrero, M.}, \bibinfo{author}{Chang, L.}, \bibinfo{author}{Pérez-Suárez, A.}, \bibinfo{author}{Busch, C.}, \bibinfo{year}{2021}.
\newblock \bibinfo{title}{Fingerprint presentation attack detection based on local features encoding for unknown attacks}.
\newblock \bibinfo{journal}{IEEE Access} \bibinfo{volume}{9}, \bibinfo{pages}{5806--5820}.
\newblock \DOIprefix\doi{10.1109/ACCESS.2020.3048756}.
%Type = Inproceedings
\bibitem[{Gragnaniello et~al.(2013)Gragnaniello, Poggi, Sansone and Verdoliva}]{Gragnaniello2013}
\bibinfo{author}{Gragnaniello, D.}, \bibinfo{author}{Poggi, G.}, \bibinfo{author}{Sansone, C.}, \bibinfo{author}{Verdoliva, L.}, \bibinfo{year}{2013}.
\newblock \bibinfo{title}{Fingerprint liveness detection based on weber local image descriptor}, in: \bibinfo{booktitle}{2013 IEEE Workshop on Biometric Measurements and Systems for Security and Medical Applications}, pp. \bibinfo{pages}{46--50}.
%Type = Article
\bibitem[{Gragnaniello et~al.(2015)Gragnaniello, Poggi, Sansone and Verdoliva}]{Gragnaniello2015}
\bibinfo{author}{Gragnaniello, D.}, \bibinfo{author}{Poggi, G.}, \bibinfo{author}{Sansone, C.}, \bibinfo{author}{Verdoliva, L.}, \bibinfo{year}{2015}.
\newblock \bibinfo{title}{Local contrast phase descriptor for fingerprint liveness detection}.
\newblock \bibinfo{journal}{Pattern Recognition} \bibinfo{volume}{48}, \bibinfo{pages}{1050--1058}.
%Type = Inproceedings
\bibitem[{Huang et~al.(2017)Huang, Liu, Maaten and Weinberger}]{b9_huang}
\bibinfo{author}{Huang, G.}, \bibinfo{author}{Liu, Z.}, \bibinfo{author}{Maaten, L.V.D.}, \bibinfo{author}{Weinberger, K.Q.}, \bibinfo{year}{2017}.
\newblock \bibinfo{title}{Densely connected convolutional networks}, in: \bibinfo{booktitle}{2017 IEEE Conference on Computer Vision and Pattern Recognition (CVPR)}, \bibinfo{publisher}{IEEE Computer Society}, \bibinfo{address}{Los Alamitos, CA, USA}. pp. \bibinfo{pages}{2261--2269}.
%Type = Article
\bibitem[{Jing~Li(2024)}]{JingLi}
\bibinfo{author}{Jing~Li, Yang~Wang, E.Z.}, \bibinfo{year}{2024}.
\newblock \bibinfo{title}{Striver: an image descriptor for fingerprint liveness detection}.
\newblock \bibinfo{journal}{Signal Image Video Process.} \bibinfo{volume}{18}, \bibinfo{pages}{8229--8239}.
\newblock \URLprefix \url{https://api.semanticscholar.org/CorpusID:271763580}.
%Type = Article
\bibitem[{Jung and Heo(2018)}]{b47_jung2}
\bibinfo{author}{Jung, H.Y.}, \bibinfo{author}{Heo, Y.}, \bibinfo{year}{2018}.
\newblock \bibinfo{title}{Fingerprint liveness map construction using convolutional neural network}.
\newblock \bibinfo{journal}{Electronics Letters} \bibinfo{volume}{54}.
%Type = Article
\bibitem[{Jung et~al.(2019)Jung, Heo and Lee}]{b45_jung}
\bibinfo{author}{Jung, H.Y.}, \bibinfo{author}{Heo, Y.S.}, \bibinfo{author}{Lee, S.}, \bibinfo{year}{2019}.
\newblock \bibinfo{title}{Fingerprint liveness detection by a template-probe convolutional neural network}.
\newblock \bibinfo{journal}{IEEE Access} \bibinfo{volume}{7}, \bibinfo{pages}{118986--118993}.
%Type = Article
\bibitem[{Kaur(2024)}]{BineetKaur}
\bibinfo{author}{Kaur, B.}, \bibinfo{year}{2024}.
\newblock \bibinfo{title}{Fingerprint and iris liveness detection using invariant feature-set}.
\newblock \bibinfo{journal}{Multimedia Tools and Applications} , \bibinfo{pages}{60833--60859}\URLprefix \url{https://doi.org/10.1007/s11042-023-17854-w}.
%Type = Article
\bibitem[{Kim(2017)}]{b23_kim}
\bibinfo{author}{Kim, W.}, \bibinfo{year}{2017}.
\newblock \bibinfo{title}{Fingerprint liveness detection using local coherence patterns}.
\newblock \bibinfo{journal}{IEEE Signal Processing Letters} \bibinfo{volume}{24}, \bibinfo{pages}{51--55}.
%Type = Article
\bibitem[{Krizhevsky et~al.(2012)Krizhevsky, Sutskever and Hinton}]{Alexnet2012}
\bibinfo{author}{Krizhevsky, A.}, \bibinfo{author}{Sutskever, I.}, \bibinfo{author}{Hinton, G.}, \bibinfo{year}{2012}.
\newblock \bibinfo{title}{Imagenet classification with deep convolutional neural networks}.
\newblock \bibinfo{journal}{Neural Information Processing Systems} \bibinfo{volume}{25}.
%Type = Article
\bibitem[{Manivanan et~al.(2010)Manivanan, Memon and Balachandran}]{Manivanan2010}
\bibinfo{author}{Manivanan, N.}, \bibinfo{author}{Memon, S.}, \bibinfo{author}{Balachandran, W.}, \bibinfo{year}{2010}.
\newblock \bibinfo{title}{Automatic detection of active sweat pores of fingerprint using highpass and correlation filtering}.
\newblock \bibinfo{journal}{Electronics letters} \bibinfo{volume}{46}, \bibinfo{pages}{1}.
%Type = Inproceedings
\bibitem[{Marcialis et~al.(2010)Marcialis, Roli and Tidu}]{b17_marcialis}
\bibinfo{author}{Marcialis, G.L.}, \bibinfo{author}{Roli, F.}, \bibinfo{author}{Tidu, A.}, \bibinfo{year}{2010}.
\newblock \bibinfo{title}{Analysis of fingerprint pores for vitality detection}, in: \bibinfo{booktitle}{2010 20th International Conference on Pattern Recognition}, pp. \bibinfo{pages}{1289--1292}.
%Type = Inproceedings
\bibitem[{Memon et~al.(2011)Memon, Manivannan and Balachandran}]{Memon2011}
\bibinfo{author}{Memon, S.}, \bibinfo{author}{Manivannan, N.}, \bibinfo{author}{Balachandran, W.}, \bibinfo{year}{2011}.
\newblock \bibinfo{title}{Active pore detection for liveness in fingerprint identification system}, in: \bibinfo{booktitle}{2011 19thTelecommunications Forum (TELFOR) Proceedings of Papers}, \bibinfo{organization}{IEEE}. pp. \bibinfo{pages}{619--622}.
%Type = Inproceedings
\bibitem[{Mura et~al.(2015)Mura, Ghiani, Marcialis, Roli, Yambay and Schuckers}]{b34_livdet2015}
\bibinfo{author}{Mura, V.}, \bibinfo{author}{Ghiani, L.}, \bibinfo{author}{Marcialis, G.L.}, \bibinfo{author}{Roli, F.}, \bibinfo{author}{Yambay, D.A.}, \bibinfo{author}{Schuckers, S.A.}, \bibinfo{year}{2015}.
\newblock \bibinfo{title}{Livdet 2015 fingerprint liveness detection competition 2015}, in: \bibinfo{booktitle}{2015 IEEE 7th International Conference on Biometrics Theory, Applications and Systems (BTAS)}, pp. \bibinfo{pages}{1--6}.
%Type = Article
\bibitem[{Nagaty(2001)}]{Finger_NN}
\bibinfo{author}{Nagaty, K.A.}, \bibinfo{year}{2001}.
\newblock \bibinfo{title}{Fingerprints classification using artificial neural networks: a combined structural and statistical approach}.
\newblock \bibinfo{journal}{Neural Networks} \bibinfo{volume}{14}, \bibinfo{pages}{1293--1305}.
\newblock \URLprefix \url{https://www.sciencedirect.com/science/article/pii/S0893608001000867}, \DOIprefix\doi{https://doi.org/10.1016/S0893-6080(01)00086-7}.
%Type = Article
\bibitem[{Nguyen et~al.(2018a)Nguyen, Baek, Pham and Park}]{Nguyen2}
\bibinfo{author}{Nguyen, D.T.}, \bibinfo{author}{Baek, N.R.}, \bibinfo{author}{Pham, T.D.}, \bibinfo{author}{Park, K.R.}, \bibinfo{year}{2018}a.
\newblock \bibinfo{title}{Presentation attack detection for iris recognition system using nir camera sensor}.
\newblock \bibinfo{journal}{Sensors} \bibinfo{volume}{18}.
\newblock \URLprefix \url{https://www.mdpi.com/1424-8220/18/5/1315}, \DOIprefix\doi{10.3390/s18051315}.
%Type = Article
\bibitem[{Nguyen et~al.(2018b)Nguyen, Pham, Baek and Park}]{Nguyen}
\bibinfo{author}{Nguyen, D.T.}, \bibinfo{author}{Pham, T.D.}, \bibinfo{author}{Baek, N.R.}, \bibinfo{author}{Park, K.R.}, \bibinfo{year}{2018}b.
\newblock \bibinfo{title}{Combining deep and handcrafted image features for presentation attack detection in face recognition systems using visible-light camera sensors}.
\newblock \bibinfo{journal}{Sensors} \bibinfo{volume}{18}.
\newblock \URLprefix \url{https://www.mdpi.com/1424-8220/18/3/699}, \DOIprefix\doi{10.3390/s18030699}.
%Type = Article
\bibitem[{Nogueira et~al.(2016)Nogueira, de~Alencar~Lotufo and Campos~Machado}]{Nogueira2016}
\bibinfo{author}{Nogueira, R.F.}, \bibinfo{author}{de~Alencar~Lotufo, R.}, \bibinfo{author}{Campos~Machado, R.}, \bibinfo{year}{2016}.
\newblock \bibinfo{title}{Fingerprint liveness detection using convolutional neural networks}.
\newblock \bibinfo{journal}{IEEE Transactions on Information Forensics and Security} \bibinfo{volume}{11}, \bibinfo{pages}{1206--1213}.
%Type = Inproceedings
\bibitem[{Orrù et~al.(2019)Orrù, Casula, Tuveri, Bazzoni, Dessalvi, Micheletto, Ghiani and Marcialis}]{b30_orru}
\bibinfo{author}{Orrù, G.}, \bibinfo{author}{Casula, R.}, \bibinfo{author}{Tuveri, P.}, \bibinfo{author}{Bazzoni, C.}, \bibinfo{author}{Dessalvi, G.}, \bibinfo{author}{Micheletto, M.}, \bibinfo{author}{Ghiani, L.}, \bibinfo{author}{Marcialis, G.}, \bibinfo{year}{2019}.
\newblock \bibinfo{title}{Livdet in action - fingerprint liveness detection competition 2019}, pp. \bibinfo{pages}{1--6}.
%Type = Inproceedings
\bibitem[{Park et~al.(2016)Park, Kim, Li, Kim and Kim}]{Park2016}
\bibinfo{author}{Park, E.}, \bibinfo{author}{Kim, W.}, \bibinfo{author}{Li, Q.}, \bibinfo{author}{Kim, J.}, \bibinfo{author}{Kim, H.}, \bibinfo{year}{2016}.
\newblock \bibinfo{title}{Fingerprint liveness detection using cnn features of random sample patches}, in: \bibinfo{booktitle}{2016 International Conference of the Biometrics Special Interest Group (BIOSIG)}, pp. \bibinfo{pages}{1--4}.
%Type = Article
\bibitem[{park et~al.(2018)park, Jang and Lee}]{b11}
\bibinfo{author}{park, Y.}, \bibinfo{author}{Jang, U.}, \bibinfo{author}{Lee, E.C.}, \bibinfo{year}{2018}.
\newblock \bibinfo{title}{Statistical anti-spoofing method for fingerprint recognition}.
\newblock \bibinfo{journal}{Soft Computing} \bibinfo{volume}{22}, \bibinfo{pages}{1--7}.
%Type = Article
\bibitem[{Rai et~al.(2024)Rai, Anshul, Jha, Jain, Sharma and Dey}]{anuj_1}
\bibinfo{author}{Rai, A.}, \bibinfo{author}{Anshul, A.}, \bibinfo{author}{Jha, A.}, \bibinfo{author}{Jain, P.}, \bibinfo{author}{Sharma, R.P.}, \bibinfo{author}{Dey, S.}, \bibinfo{year}{2024}.
\newblock \bibinfo{title}{An open patch generator based fingerprint presentation attack detection using generative adversarial network}.
\newblock \bibinfo{journal}{Multimedia Tools and Applications} \bibinfo{volume}{83}, \bibinfo{pages}{27723--27746}.
%Type = Inproceedings
\bibitem[{Rai and Dey(2024)}]{anuj_5}
\bibinfo{author}{Rai, A.}, \bibinfo{author}{Dey, S.}, \bibinfo{year}{2024}.
\newblock \bibinfo{title}{An explainable deep learning model for fingerprint presentation attack detection}, in: \bibinfo{editor}{Kaur, H.}, \bibinfo{editor}{Jakhetiya, V.}, \bibinfo{editor}{Goyal, P.}, \bibinfo{editor}{Khanna, P.}, \bibinfo{editor}{Raman, B.}, \bibinfo{editor}{Kumar, S.} (Eds.), \bibinfo{booktitle}{Computer Vision and Image Processing}, \bibinfo{publisher}{Springer Nature Switzerland}, \bibinfo{address}{Cham}. pp. \bibinfo{pages}{309--321}.

%Type = Article
\bibitem[{Rai et~al.(2025)Rai, Dey, Patidar and Rai}]{anuj_4}
\bibinfo{author}{Rai, A.}, \bibinfo{author}{Dey, S.}, \bibinfo{author}{Patidar, P.}, \bibinfo{author}{Rai, P.}, \bibinfo{year}{2025}.
\newblock \bibinfo{title}{Mosfpad: An end-to-end ensemble of mobilenet and support vector classifier for fingerprint presentation attack detection}.
\newblock \bibinfo{journal}{Computers and Security} \bibinfo{volume}{148}, \bibinfo{pages}{104069}.
\newblock \URLprefix \url{https://www.sciencedirect.com/science/article/pii/S0167404824003742}, \DOIprefix\doi{https://doi.org/10.1016/j.cose.2024.104069}.

%Type = Inproceedings
\bibitem[{Rattani and Ross(2014)}]{Rattani2014_1}
\bibinfo{author}{Rattani, A.}, \bibinfo{author}{Ross, A.}, \bibinfo{year}{2014}.
\newblock \bibinfo{title}{Automatic adaptation of fingerprint liveness detector to new spoof materials}, in: \bibinfo{booktitle}{IEEE International Joint Conference on Biometrics}, pp. \bibinfo{pages}{1--8}.
%Type = Article
\bibitem[{Rattani et~al.(2015)Rattani, Scheirer and Ross}]{Rattani2015}
\bibinfo{author}{Rattani, A.}, \bibinfo{author}{Scheirer, W.J.}, \bibinfo{author}{Ross, A.}, \bibinfo{year}{2015}.
\newblock \bibinfo{title}{Open set fingerprint spoof detection across novel fabrication materials}.
\newblock \bibinfo{journal}{IEEE Transactions on Information Forensics and Security} \bibinfo{volume}{10}, \bibinfo{pages}{2447--2460}.
%Type = Article
\bibitem[{Rubab~Mehboob(2023)}]{Rubab}
\bibinfo{author}{Rubab~Mehboob, Hassan~Dawood, H.D.}, \bibinfo{year}{2023}.
\newblock \bibinfo{title}{An encoded histogram of ridge bifurcations and contours for fingerprint presentation attack detection}.
\newblock \bibinfo{journal}{Pattern Recognition} \bibinfo{volume}{143}, \bibinfo{pages}{109782}.
\newblock \URLprefix \url{https://www.sciencedirect.com/science/article/pii/S0031320323004806}, \DOIprefix\doi{https://doi.org/10.1016/j.patcog.2023.109782}.
%Type = Article
\bibitem[{Sharma and Selwal(2021)}]{b37_deepika}
\bibinfo{author}{Sharma, D.}, \bibinfo{author}{Selwal, A.}, \bibinfo{year}{2021}.
\newblock \bibinfo{title}{Hyfipad: a hybrid approach for fingerprint presentation attack detection using local and adaptive image features}.
\newblock \bibinfo{journal}{The Visual Computer} .
%Type = Article
\bibitem[{Sharma and Dey(2019)}]{b5_sharma1}
\bibinfo{author}{Sharma, R.}, \bibinfo{author}{Dey, S.}, \bibinfo{year}{2019}.
\newblock \bibinfo{title}{Fingerprint liveness detection using local quality features}.
\newblock \bibinfo{journal}{The Visual Computer} \bibinfo{volume}{35}.
%Type = Inproceedings
\bibitem[{Toosi. et~al.(2017)Toosi., Cumani. and Bottino.}]{Toosi2017}
\bibinfo{author}{Toosi., A.}, \bibinfo{author}{Cumani., S.}, \bibinfo{author}{Bottino., A.}, \bibinfo{year}{2017}.
\newblock \bibinfo{title}{Cnn patch--based voting for fingerprint liveness detection}, in: \bibinfo{booktitle}{Proceedings of the 9th International Joint Conference on Computational Intelligence (IJCCI 2017) - IJCCI}, \bibinfo{organization}{INSTICC}. \bibinfo{publisher}{SciTePress}. pp. \bibinfo{pages}{158--165}.
%Type = Article
\bibitem[{Uliyan et~al.(2020)Uliyan, Sadeghi and Jalab}]{b4_uliyan}
\bibinfo{author}{Uliyan, D.M.}, \bibinfo{author}{Sadeghi, S.}, \bibinfo{author}{Jalab, H.A.}, \bibinfo{year}{2020}.
\newblock \bibinfo{title}{Anti-spoofing method for fingerprint recognition using patch based deep learning machine}.
\newblock \bibinfo{journal}{Engineering Science and Technology, an International Journal} \bibinfo{volume}{23}, \bibinfo{pages}{264--273}.
%Type = Article
\bibitem[{Wu et~al.(2020)Wu, Li, Zhao and Liu}]{cifar}
\bibinfo{author}{Wu, C.}, \bibinfo{author}{Li, Y.}, \bibinfo{author}{Zhao, Z.}, \bibinfo{author}{Liu, B.}, \bibinfo{year}{2020}.
\newblock \bibinfo{title}{Research on image classification method of features of combinatorial convolution}.
\newblock \bibinfo{journal}{Journal of Ambient Intelligence and Humanized Computing} \bibinfo{volume}{11}.
%Type = Article
\bibitem[{Xia et~al.(2020)Xia, Yuan, Lv, Sun, Xiong and Shi}]{xia_2}
\bibinfo{author}{Xia, Z.}, \bibinfo{author}{Yuan, C.}, \bibinfo{author}{Lv, R.}, \bibinfo{author}{Sun, X.}, \bibinfo{author}{Xiong, N.N.}, \bibinfo{author}{Shi, Y.Q.}, \bibinfo{year}{2020}.
\newblock \bibinfo{title}{A novel weber local binary descriptor for fingerprint liveness detection}.
\newblock \bibinfo{journal}{IEEE Transactions on Systems, Man, and Cybernetics: Systems} \bibinfo{volume}{50}, \bibinfo{pages}{1526--1536}.
%Type = Inproceedings
\bibitem[{Yadav et~al.(2018)Yadav, Kohli, Agarwal, Vatsa, Singh and Noore}]{Daksha_Yadav}
\bibinfo{author}{Yadav, D.}, \bibinfo{author}{Kohli, N.}, \bibinfo{author}{Agarwal, A.}, \bibinfo{author}{Vatsa, M.}, \bibinfo{author}{Singh, R.}, \bibinfo{author}{Noore, A.}, \bibinfo{year}{2018}.
\newblock \bibinfo{title}{Fusion of handcrafted and deep learning features for large-scale multiple iris presentation attack detection}, in: \bibinfo{booktitle}{2018 IEEE/CVF Conference on Computer Vision and Pattern Recognition Workshops (CVPRW)}, pp. \bibinfo{pages}{685--6857}.
\newblock \DOIprefix\doi{10.1109/CVPRW.2018.00099}.
%Type = Inproceedings
\bibitem[{Yambay et~al.(2018)Yambay, Schuckers, Denning, Sandmann, Bachurinski and Hogan}]{livdet_2017}
\bibinfo{author}{Yambay, D.}, \bibinfo{author}{Schuckers, S.}, \bibinfo{author}{Denning, S.}, \bibinfo{author}{Sandmann, C.}, \bibinfo{author}{Bachurinski, A.}, \bibinfo{author}{Hogan, J.}, \bibinfo{year}{2018}.
\newblock \bibinfo{title}{Livdet 2017 - fingerprint systems liveness detection competition}, in: \bibinfo{booktitle}{2018 IEEE 9th International Conference on Biometrics Theory, Applications and Systems (BTAS)}, pp. \bibinfo{pages}{1--9}.
%Type = Article
\bibitem[{Yuan et~al.(2022)Yuan, Jiao, Sun and Wu}]{Yuan2022}
\bibinfo{author}{Yuan, C.}, \bibinfo{author}{Jiao, S.}, \bibinfo{author}{Sun, X.}, \bibinfo{author}{Wu, Q.M.J.}, \bibinfo{year}{2022}.
\newblock \bibinfo{title}{Mfffld: A multimodal-feature-fusion-based fingerprint liveness detection}.
\newblock \bibinfo{journal}{IEEE Transactions on Cognitive and Developmental Systems} \bibinfo{volume}{14}, \bibinfo{pages}{648--661}.
%Type = Article
\bibitem[{Yuan et~al.(2019)Yuan, Xia, Jiang, Cao, Jonathan~Wu and Sun}]{b44_yuan3}
\bibinfo{author}{Yuan, C.}, \bibinfo{author}{Xia, Z.}, \bibinfo{author}{Jiang, L.}, \bibinfo{author}{Cao, Y.}, \bibinfo{author}{Jonathan~Wu, Q.M.}, \bibinfo{author}{Sun, X.}, \bibinfo{year}{2019}.
\newblock \bibinfo{title}{Fingerprint liveness detection using an improved cnn with image scale equalization}.
\newblock \bibinfo{journal}{IEEE Access} \bibinfo{volume}{7}, \bibinfo{pages}{26953--26966}.
%Type = Article
\bibitem[{Yuan et~al.(2020)Yuan, Xia, Sun and Wu}]{Yuan2020}
\bibinfo{author}{Yuan, C.}, \bibinfo{author}{Xia, Z.}, \bibinfo{author}{Sun, X.}, \bibinfo{author}{Wu, Q.M.J.}, \bibinfo{year}{2020}.
\newblock \bibinfo{title}{Deep residual network with adaptive learning framework for fingerprint liveness detection}.
\newblock \bibinfo{journal}{IEEE Transactions on Cognitive and Developmental Systems} \bibinfo{volume}{12}, \bibinfo{pages}{461--473}.
%Type = Article
\bibitem[{Zhang et~al.(2019)Zhang, Shi, Zhan, Cao, Zhu and Li}]{b31_zhang}
\bibinfo{author}{Zhang, Y.}, \bibinfo{author}{Shi, D.}, \bibinfo{author}{Zhan, X.}, \bibinfo{author}{Cao, D.}, \bibinfo{author}{Zhu, K.}, \bibinfo{author}{Li, Z.}, \bibinfo{year}{2019}.
\newblock \bibinfo{title}{Slim-rescnn: A deep residual convolutional neural network for fingerprint liveness detection}.
\newblock \bibinfo{journal}{IEEE Access} \bibinfo{volume}{7}, \bibinfo{pages}{91476--91487}.
%Type = Article
\bibitem[{Šmida et~al.(2015)Šmida, Busch and Olsen}]{olsen}
\bibinfo{author}{Šmida, V.}, \bibinfo{author}{Busch, C.}, \bibinfo{author}{Olsen, M.}, \bibinfo{year}{2015}.
\newblock \bibinfo{title}{Finger image quality assessment features – definitions and evaluation}.
\newblock \bibinfo{journal}{IET Biometrics} \bibinfo{volume}{5}.

\end{thebibliography}

%\vskip10pt

\end{document}